\documentclass{article} 
\PassOptionsToPackage{dvipsnames,table}{xcolor}
\usepackage{iclr2027_conference,times}

\usepackage{amsmath,amsfonts,bm}

\def\eqref#1{equation~\ref{#1}}

\def\1{\bm{1}}

\DeclareMathAlphabet{\mathsfit}{\encodingdefault}{\sfdefault}{m}{sl}
\SetMathAlphabet{\mathsfit}{bold}{\encodingdefault}{\sfdefault}{bx}{n}

\usepackage{url}
\usepackage{booktabs}
\usepackage{graphicx}
\graphicspath{{figures/}}
\usepackage{multirow}
\usepackage{xcolor}
\definecolor{rankone}{HTML}{D6E7F8}   
\definecolor{ranktwo}{HTML}{D8EEDC}   
\definecolor{rankthree}{HTML}{F8EBCB} 

\newcommand{\cellone}[1]{\cellcolor{rankone}\textbf{#1}}
\newcommand{\celltwo}[1]{\cellcolor{ranktwo}#1}
\newcommand{\cellthree}[1]{\cellcolor{rankthree}#1}

\newcommand{\swone}[1]{\colorbox{rankone}{#1}}
\newcommand{\swtwo}[1]{\colorbox{ranktwo}{#1}}
\newcommand{\swthree}[1]{\colorbox{rankthree}{#1}}

\usepackage{amsmath}
\usepackage{amssymb}
\usepackage{algorithm}
\usepackage{algpseudocode}
\usepackage{longtable}
\usepackage{array}

\definecolor{myiclrblue}{RGB}{152, 182, 231}
\usepackage[colorlinks=true,
            linkcolor=myiclrblue,
            citecolor=myiclrblue,
            urlcolor=myiclrblue]{hyperref}
\usepackage{cleveref}

\newcommand{\navsafe}{\texorpdfstring{\textsc{NavSafe-$\infty$}}{NavSafe-∞}}

\newcommand{\nurec}{\textsc{NuRec}}

\newcommand{\navsafelogo}{%
    \raisebox{-0.28\height}{\includegraphics[height=2.6ex]{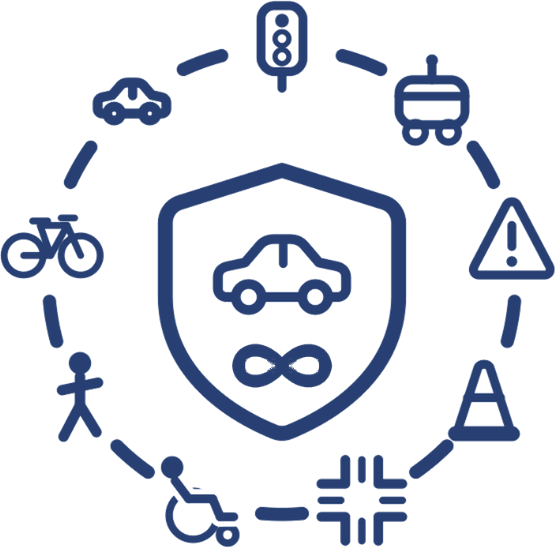}}}

\usepackage{subcaption}

\usepackage{array}
\usepackage{longtable}
\usepackage{booktabs}

\usepackage{seqsplit}

\definecolor{Terracotta}{HTML}{D2691E}

\title{\navsafelogo\hspace{0.35em}\navsafe{}: Benchmarking Closed-Loop Driving Safety in Photorealistic Environments}

\author{Yuxin Bao$^1$\thanks{Equal Contribution; Order by Last Name.} ~~~~~
Hongwei Ruan$^2$\footnotemark[1]~~~~~
Luobin Wang$^2$\thanks{Corresponding authors: \texttt{luw015@ucsd.edu} and \texttt{sethzhao506@g.ucla.edu}.}~~~~~
Seth Z. Zhao$^1$\footnotemark[2] ~~~~~
Ziyang Leng$^1$ \And
Zihan Zhang$^2$~~~~~
Yu Zeng$^3$~~~~~
Rowan McAllister$^3$~~~~~
Henrik Christensen$^2$~~~~~
Bolei Zhou$^1$ \\
$^1$UCLA~~~~~$^2$UCSD~~~~~$^3$Toyota Research Institute \\
\tt{\href{https://navsafe-vail.github.io/}{\tt https://navsafe-vail.github.io/}}
}

\iclrfinalcopy 

\begin{document}
\maketitle
\begin{figure}[tbh]
  \centering
  \includegraphics[width=\textwidth]{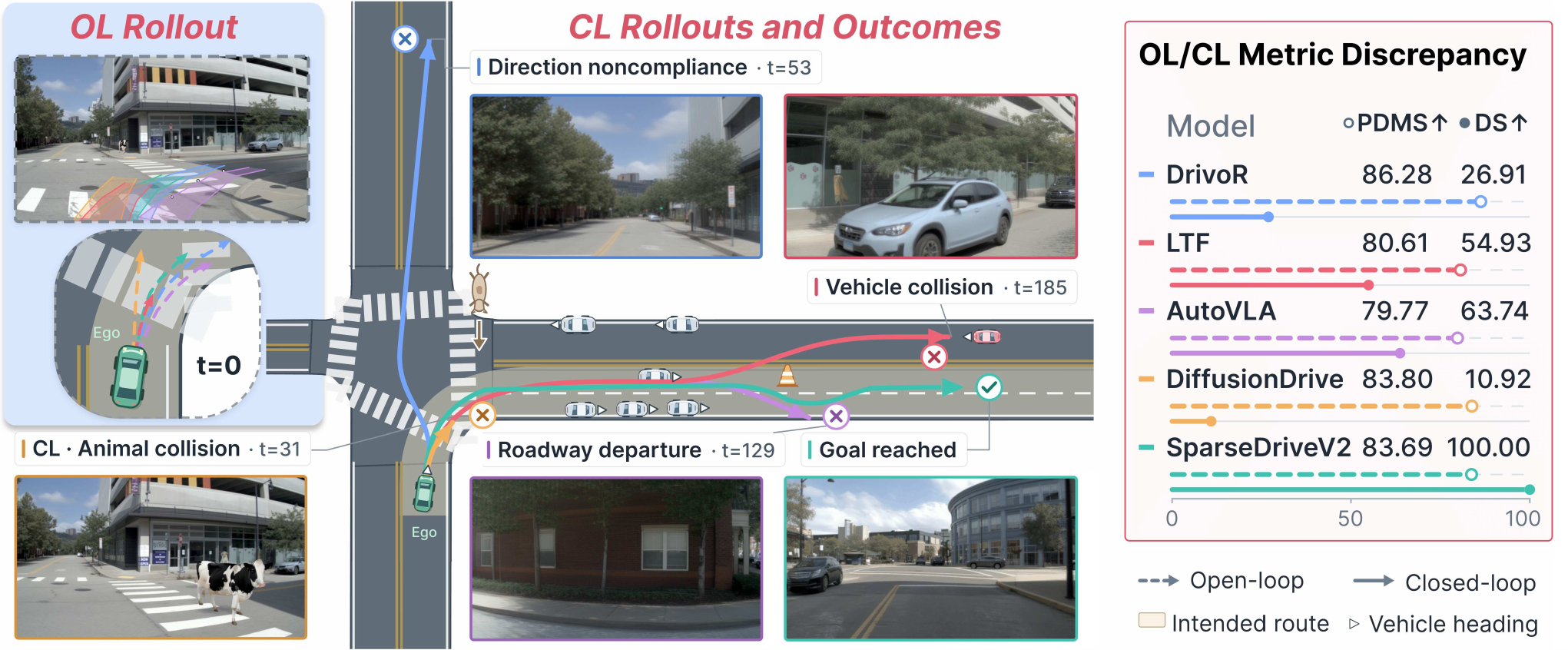}
  \caption{\textbf{Motivations.} End-to-end (E2E) driving policies with comparable open-loop (OL) predictive driver model scores (PDMS) diverge under closed-loop (CL) execution, yielding different safety outcomes and Driving Scores (DS), which reveals the OL-to-CL evaluation gap in E2E driving.}
  \label{fig:motivation}
\end{figure}

\begin{abstract}
End-to-end (E2E) driving policies have advanced rapidly on open-loop (OL) benchmarks, yet OL evaluation cannot reveal whether a policy can withstand compounding errors, recover from failures, or interact safely with surrounding actors. We introduce \navsafe{}, a photorealistic closed-loop (CL) benchmark comprising 280 scenarios spanning 28 event types, each with success and failure criteria defined within a structured traffic-safety taxonomy, yielding category-level capability scores for Traffic Crashes, Vulnerable Road User Crashes, Traffic Violations, and Traffic Incidents. After evaluating 20 E2E policies, we find that OL gains do not reliably transfer to CL safety. Analysis of two common remedies reveals that (1) passive demonstration perturbation helps mainly when CL rollouts stay near their perturbed training states, and (2) OL reinforcement-learning fine-tuning exhibits reward hacking by trading safety margin for ego progress, which CL feedback amplifies into compounding safety-critical errors. Together, these results demonstrate the blind spot of OL benchmarks indicating CL safety success. The benchmark and an extensible toolbox for customizable event curation and policy diagnosis will be open-sourced to facilitate future research.
\end{abstract}

\section{Introduction}
Real-world driving safety depends on how a policy resolves safety-critical situations through continuous interaction: in closed-loop (CL) driving, a policy's actions reshape its future observations and the surrounding dynamics, so the same scenario can lead to divergent outcomes. Open-loop (OL) evaluation on static logged states cannot capture these interaction sequences or the compounding errors they cause. Yet recent end-to-end (E2E) driving policies are evaluated primarily on OL benchmarks such as nuScenes and NAVSIM~\citep{Caesar2020nuScenes,Dauner2024NAVSIM,Cao2025PseudoSimulation}, where safety metrics score predicted trajectories without sequential policy–environment feedback. Figure~\ref{fig:motivation} illustrates this OL-to-CL evaluation gap: policies with similar OL scores yield different CL safety outcomes in the same scenario. Thus, the community still has limited insight into the reliability of current E2E policies confronting safety-critical events.

In this paper, we aim to understand whether existing driving policies can reliably navigate long-horizon CL scenarios across sequences of safety-relevant events. To systematically evaluate these capabilities across diverse, safety-relevant interactions, we introduce \navsafe{}, a photorealistic CL benchmark that evaluates policies against event-specific objectives derived from real-world traffic-safety events~\citep{RoadSafe365}.  

Each event specifies success and failure conditions that are evaluated over the executed CL rollout. For example, a red-light event requires the ego to remain behind the stop line for the prescribed interval, recognizing compliant stopping as successful task completion rather than route completion. Through event-level evaluation, \navsafe{} enables identification of specific safety failures and assessment of whether gains in OL performance translate into improved CL safety.

To assess policy behavior in CL, we conduct a fine-grained analysis across the \navsafe{} safety taxonomy. Our results show that strong OL performance does not reliably transfer to CL safety: high-ranking policies on OL benchmarks remain vulnerable across multiple safety-critical event types when operating in policy-induced states. We further examine whether common mitigation practices for narrowing the OL-to-CL gap, such as passive demonstration perturbation~\citep{Karkus2025BeyondBC, Tian2026SimScale} and reinforcement learning (RL) fine-tuning (RLFT) methods, alleviate these failures.  Notably, we find that 1) perturbation (training on states perturbed independently of the policy) helps mainly when CL rollouts stay close to those states and 2) RLFT exhibits signs of reward hacking in the OL PDM score (PDMS)~\citep{Dauner2024NAVSIM}, while CL feedback can amplify the resulting errors and degrade CL performance.  These findings shed light on how future methods should be developed to improve CL driving safety.

Our contributions are threefold. \textbf{1) \navsafe{} Benchmark:} we present a photorealistic CL benchmark that evaluates driving safety at the event level, grounded in a real-world traffic-safety taxonomy, and validate its rendering fidelity with perceptual and policy-oriented metrics so that policy behavior is evaluated with limited visual distortion (Section~\ref{sec:rendering_fidelity}). \textbf{2) Analysis:} we show that OL performance does not reliably translate into CL safety, localize the resulting failure patterns across policies, and diagnose why perturbation and RLFT yield inconsistent CL gains. \textbf{3) \navsafe{} Toolbox:} we design an extensible toolbox for customizable scenario curation and policy diagnosis under controlled changes to ego and environment states, enabling the community to extend \navsafe{} with new safety-critical events and diagnostic protocols. Together, the benchmark and toolbox provide a reliable and extensible testbed for studying various CL driving behaviors.

\section{Related Work}
\subsection{End-to-End Driving Benchmarks}
\textbf{Open-Loop Benchmarks.} OL benchmarks offer a low-cost, scalable proxy on real sensor observations. nuScenes~\citep{Caesar2020nuScenes} scores trajectories against recorded futures; NAVSIM~\citep{Dauner2024NAVSIM,Cao2025PseudoSimulation} replaces displacement error with collision- and compliance-based planning scores; WOMD-Reasoning~\citep{li2024womd} and nuReasoning~\citep{Huang2026nuReasoning} add reasoning question-answer pairs. Yet the policy never sees the consequences of its own actions, hiding compounding error, failure recovery, and reactive traffic, and OL rankings only loosely track CL performance~\citep{Zhao2026BridgeSim,Wang2026OLCLCorrelation}.

\textbf{Closed-Loop Benchmarks.} CL benchmarks execute policies through sequential feedback. CARLA-based~\citep{Dosovitskiy2017CARLA} Bench2Drive~\citep{Jia2024Bench2Drive}, HiDrive~\citep{xia2026hidrive}, and Fail2Drive~\citep{Gerstenecker2026Fail2Drive} provide structured behavioral evaluation, with the first two reporting per-ability results, but game-engine rendering leaves a domain gap for real-sensor policies. HUGSIM~\citep{Zhou2026HUGSIM}, WorldEngine~\citep{Li2026WorldEngine}, and AlpaSim~\citep{NVIDIA2025AlpaSim} narrow this gap with 3DGS reconstruction~\citep{Kerbl2023GaussianSplatting,NVIDIA2026NuRec}, and DriveArena~\citep{Yang2025DriveArena} and AlpaSim with generative rendering (OmniDreams~\citep{NVIDIA2026OmniDreams}), yet they mainly report aggregate scores on their own scenarios. Moreover, existing ability categories follow driving maneuvers rather than real-world safety events, and success is typically generic route completion. \navsafe{} instead combines photorealistic CL with taxonomy-grounded events, event-specific success criteria, and category-level Ability Scores.

\subsection{End-to-End Driving Policies}
\label{sec:related_policies}
We introduce the following driving policy training paradigms that are covered in this paper: 

\textbf{Imitation learning-based pretraining.} Most E2E and vision--language--action (VLA) policies are trained offline through imitation learning (IL) or supervised fine-tuning (SFT): LTF~\citep{Dauner2024NAVSIM}, DiffusionDrive~\citep{Liao2025DiffusionDrive}, and DrivoR~\citep{Kirby2026DrivoR} learn trajectories from fixed demonstrations; RAP~\citep{Feng2026RAP} augments them with rasterized counterfactual views; SparseDriveV2~\citep{Sun2026SparseDriveV2} and GTRS~\citep{Li2025GTRS} score a vocabulary factorized into paths and velocity profiles; VLA policies align vision, language, and action \citep{Zhou2025AutoVLA}. All inherit behavior cloning's covariate shift: training never covers the state distribution the policy induces at test time.

\textbf{Reinforcement learning-based post-training.} RL optimizes a driving reward atop an imitation-pretrained policy: ReCogDrive~\citep{Li2025ReCogDrive}, MTDrive~\citep{Li2026MTDrive}, and AutoVLA~\citep{Zhou2025AutoVLA} do so over offline observations. Staying offline confines reward to logged states and leaves reward-exploiting policies uncorrected by the environment.

\textbf{Passive demonstration perturbation methods.} To synthesize states absent from logs, SimScale~\citep{Tian2026SimScale} perturbs ego trajectories in reconstructed reactive scenes and co-trains with pseudo-expert supervision.  BeyondDrive~\citep{Wang2026BeyondDrive} mines expert-proximate unsafe trajectories as negatives.

\textbf{World-model-based policies.} These policies learn future prediction alongside planning: DriveLaW~\citep{Xia2026DriveLaW} conditions a flow-matching planner on video-diffusion latents, DriveVLA--W0~\citep{Li2025DriveVLAW0} adds future-image prediction as auxiliary supervision to a VLA policy, and SimWAM~\citep{Zhao2026SimWAM} co-trains an action expert with a video expert dropped at inference.

\begin{figure*}[tbh]
  \centering
  \includegraphics[width=\textwidth]{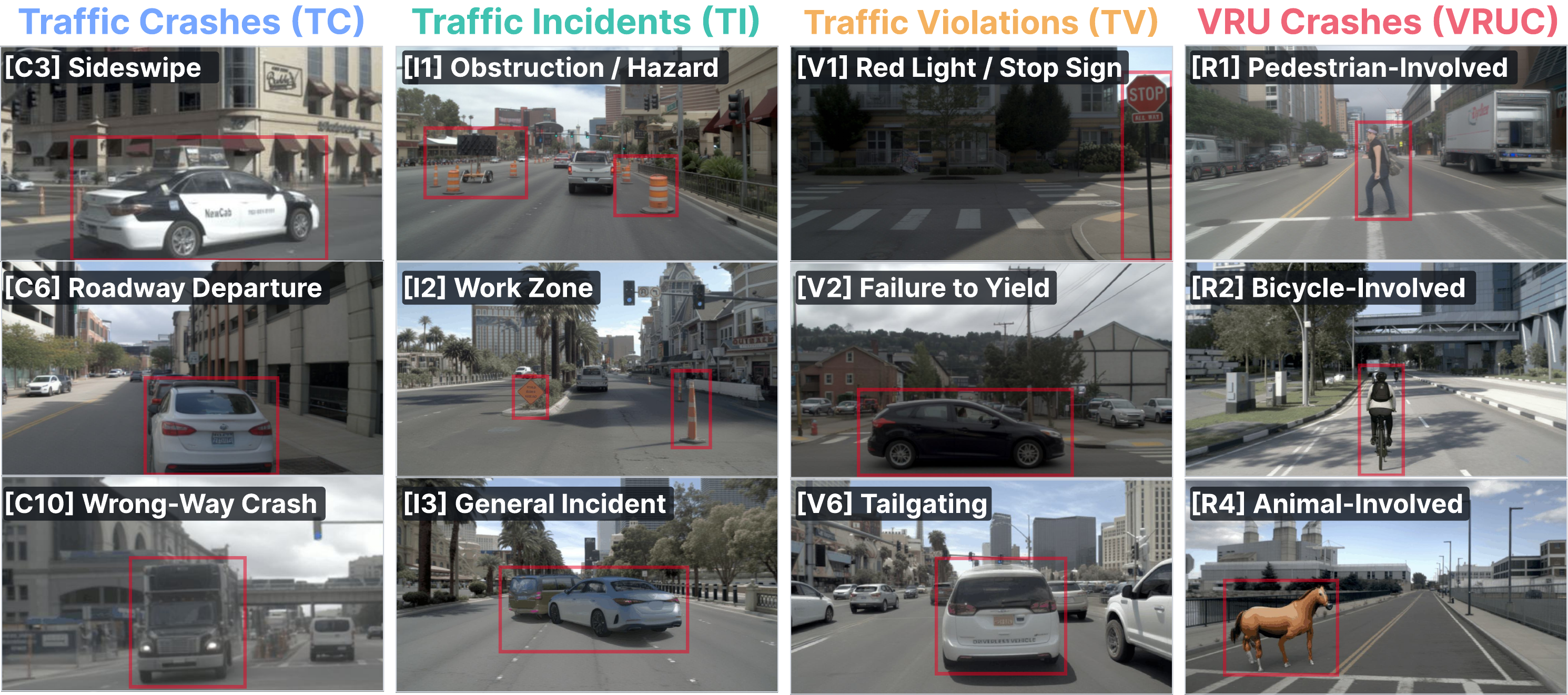}
  \caption{Representative \navsafe{} events. Appendix~\ref{app:scenario_gallery} shows the remaining 16 event types.}
  \label{fig:scenario_demo}
\end{figure*}

\section{\navsafe{} Benchmark}
\label{sec:benchmark}

\navsafe{} is a photorealistic CL benchmark comprising 280 event-based scenarios organized into 28 event types adapted from RoadSafe365~\citep{RoadSafe365}. These event types span four categories: \textit{Traffic Crashes (TC)}, \textit{Vulnerable Road User Crashes (VRUC)}, \textit{Traffic Violations (TV)}, and \textit{Traffic Incidents (TI)}. Figure~\ref{fig:scenario_demo} shows representative scenarios from each category. Section~\ref{sec:event_design} presents the event-based benchmark design, Section~\ref{sec:protocol} defines the evaluation protocol and metrics, and Section~\ref{sec:rendering_fidelity} evaluates rendering fidelity and policy-relevant visual alignment.

\subsection{Event-Based Benchmark Design}
\label{sec:event_design}

\begin{figure*}[tbh]
  \centering
  \includegraphics[width=\textwidth]{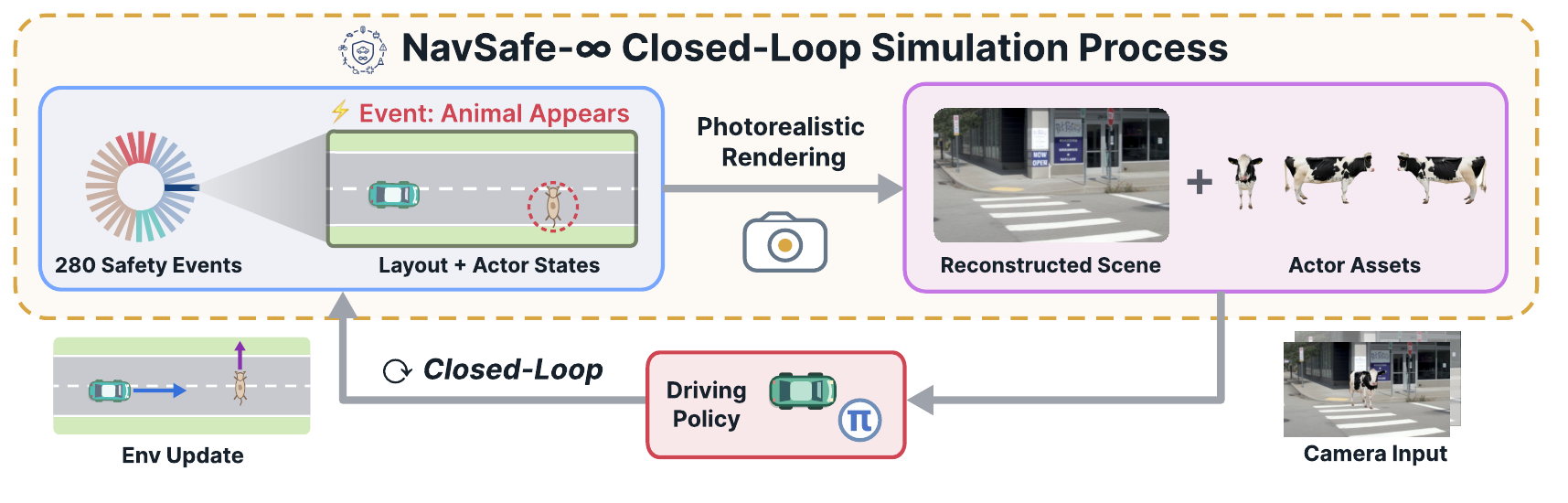}
  \caption{\textbf{\navsafe{} Closed-Loop Simulation Process.} \navsafe{} loads a bounded, event-specific reconstruction. At each step, the renderer produces camera observations from the current ego and actor states; the policy's plan advances the simulator, yielding the next observation.}
  \label{fig:curation_pipeline}
\end{figure*}

\textbf{Design Principles.} Long-horizon driving requires a policy to resolve a sequence of local interactions and traffic-rule decisions. A single route-level score aggregates behavior across these events, making it difficult to assess specific safety capabilities. Prior work further shows that long-route evaluation can produce high-variance aggregate scores~\citep{Jia2024Bench2Drive}. \navsafe{} therefore evaluates bounded, event-centered scenarios, enabling granular capability assessment while preserving CL feedback within each event. Figure~\ref{fig:curation_pipeline} illustrates how this design preserves both properties: each bounded event isolates a specific safety capability while retaining policy-environment feedback throughout the rollout.

\textbf{Event Definition.} We define an \emph{event} as a bounded traffic interaction or rule-governed situation with explicit task and safety requirements. Each scenario instantiates an event type through a specified initial scene, relevant road users or traffic controls, and termination conditions. Figure~\ref{fig:scenario_demo} demonstrates some representative examples including responding to a braking lead vehicle, yielding to a crossing pedestrian, holding behind a red-light stop line, and navigating a work-zone obstruction. The required behavior determines how success is evaluated: a red-light event requires a compliant hold for the prescribed interval, whereas a work-zone event requires completing the prescribed bypass without disallowed infractions. Appendices~\ref{app:taxonomy_coverage} and~\ref{app:event_contracts} list the inclusion criteria of all 28 event types and their success and termination rules.

\textbf{Task Setting.} Following~\citet{Zhao2026BridgeSim}, we formulate E2E CL driving as a finite-horizon partially observable Markov decision process traversing through each scenario $e$ of event type $\ell$, namely $\mathcal{M}_e=(\mathcal{S},\mathcal{A},\mathcal{O},P_e,\Omega_\kappa,\rho_e,R_\ell,H_e)$, with transition $P_e$, initial-state distribution $\rho_e$, sparse terminal success reward $R_\ell$, and horizon $H_e$. Two properties distinguish this formulation. First, the emission model $\Omega_\kappa(o_t\mid s_t)$ provides \emph{configurable camera observations}: because each scene is reconstructed as a 3DGS representation, the same scenario can be rendered as single-view, multi-view, or any other camera rig $\kappa$ by changing only the camera intrinsics and extrinsics, without re-collecting data. Second, the dynamic state $s_t$ is \emph{fully controllable}, including editing traffic elements and prescribing the behaviors of surrounding vehicles. The objective is to complete the event-specific task while maintaining safe execution throughout the CL interaction.

 \subsection{Evaluation Protocol and Metrics}
\label{sec:protocol}

Appendix~\ref{app:evaluation_protocol} provides the full evaluation protocol, while Appendix~\ref{app:reproducibility} reports runtime and resource requirements.

We report \emph{Driving Score (DS)}, \emph{Success Rate (SR)}, \emph{Driving Efficiency (DE)}, and \emph{Comfort}, following established CL driving evaluation protocols~\citep{Jia2024Bench2Drive,Caesar2021nuPlan}. For each event, DS is computed by multiplying \emph{Event Progress (EP)} by penalties for recorded infractions and scaling the result to $[0,100]$. It therefore rewards event progress while penalizing unsafe or non-compliant behavior. SR is the percentage of events that satisfy their event-type-specific success criteria within the time budget without a disallowed infraction. DE measures ego speed relative to surrounding traffic, while Comfort measures compliance with predefined bounds on acceleration, jerk, yaw rate, and yaw acceleration. We additionally report an \emph{Ability Score} for each benchmark category (TC, VRUC, TV, and TI), defined as the unweighted mean of the event-type-level SRs within that category. 

\begin{table}[tbh]
  \centering
  \caption{Quantitative visual fidelity comparison across generative renderers. All metrics compare rendered observations with the corresponding recorded clips. DD refers to DiffusionDrive~\citep{Liao2025DiffusionDrive}; SDv2 refers to SparseDriveV2~\citep{Sun2026SparseDriveV2}.}
  \label{tab:rendering_ablation}
  \setlength{\tabcolsep}{2.2pt}
  \renewcommand{\arraystretch}{1.12}
  \small
  \resizebox{0.8\textwidth}{!}{%
  \begin{tabular}{@{}lccccccccc@{}}
    \toprule
    \multirow{2}{*}{\textbf{Generative Renderer}} &
    \multirow{2}{*}{\textbf{Venue}} &
    \multirow{2}{*}{\textbf{FID$\downarrow$}} &
    \multirow{2}{*}{\textbf{FVD$\downarrow$}} &
    \multirow{2}{*}{\textbf{$\mathrm{FD}_{\pi}\downarrow$}} &
    \multicolumn{5}{c}{\textbf{$\mathrm{FD}_{\pi}^{k}\downarrow$}} \\
    \cmidrule(lr){6-10}
    & & & & &
    \textbf{DrivoR} &
    \textbf{DD} &
    \textbf{LTF} &
    \textbf{RAP} &
    \textbf{SDv2} \\
    \midrule
    DriveArena~\citep{Yang2025DriveArena} & ICCV 2025 & 27.14 & 814.81 & 18.55 & 12.88 & 33.96 & 7.61 & 27.10 & 11.18 \\
    DreamStream~\citep{leng2026dreamstream} & CoRL 2026 & 11.53 & 176.83 & 6.27 & 5.51 & 11.06 & 2.89 & 6.67 & 5.24 \\
    \navsafe{} & -- & \textbf{4.95} & \textbf{45.34} & \textbf{2.45} & \textbf{2.26} & \textbf{5.01} & \textbf{1.07} & \textbf{1.67} & \textbf{2.26} \\
    \bottomrule
  \end{tabular}}
\end{table}

\subsection{Rendering Fidelity and Policy-Relevant Visual Alignment}
\label{sec:rendering_fidelity}

Observations are rendered from 3DGS reconstructions~\citep{NVIDIA2026NuRec} and refined by the diffusion-based DiffusionHarmonizer~\citep{Zhang2026DiffusionHarmonizer} (Appendix~\ref{app:environment_details}). To assess the visual fidelity of the reconstructed observations, we compare rendered frames against the corresponding real camera frames from the source logs. We report FID and FVD~\citep{Heusel2017FID,Unterthiner2018FVD} for frame-level and video-level (temporal) distribution gaps, and the policy-oriented $\mathrm{FD}_{\pi}$~\citep{leng2026dreamstream}, which is the Fr\'{e}chet distance between policy features of real and rendered frames, averaged over five E2E policies ($\mathrm{FD}_{\pi}^{k}$ per policy; details in Appendix~\ref{app:rendering_fidelity}).

As shown in Table~\ref{tab:rendering_ablation} and Figure~\ref{fig:drivearena_qualitative}, \navsafe{} achieves better rendering quality than existing generative renderers on all three metrics.

\begin{figure}[tbh]
  \centering
  \includegraphics[width=\textwidth]{figs/fig4.pdf}
  \caption{Rendering quality on a long-horizon scenario. \textbf{Left:} First-person views at $t \in \{0, 5, 10, 15\}$s. \textbf{Right:} Top-down bird's-eye view. DriveArena shows temporal inconsistency and DreamStream exhibits cumulative drift, whereas \navsafe{} stays temporally consistent.}
  \label{fig:drivearena_qualitative}
\end{figure}

\section{Experiments}
\label{sec:experiments}

We organize our experiments around three research questions: \textbf{(1)} How closely is performance on the OL benchmark NAVSIM~\citep{Dauner2024NAVSIM} associated with success in safety-critical events under CL execution? \textbf{(2)} What failure mechanisms underlie representative unsafe CL outcomes across mainstream E2E policy paradigms? 
\textbf{(3)} Why do post-IL methods such as perturbation and RLFT yield inconsistent CL gains across policies?

\subsection{\navsafe{} Leaderboard}
\label{sec:leaderboard}

We evaluate policies using four global CL metrics and four category-specific ability scores (Table~\ref{tab:main}). Appendix~\ref{app:full_leaderboard} gives the complete leaderboard.

\begin{table*}[tbh]
      \centering
      \caption{\navsafe{} leaderboard. Among learned policies, blue, green, and yellow backgrounds denote the \swone{first-}, \swtwo{second-}, and \swthree{third-}highest point estimates in each column, respectively. $\pm$ is the sample standard deviation across the three per-seed estimates.}
      \label{tab:main}
      \setlength{\tabcolsep}{1.6pt}
      \renewcommand{\arraystretch}{1.15}
      \scriptsize
      \resizebox{\textwidth}{!}{%
      \begin{tabular}{@{}llcccccccc@{}}
      \toprule
      \multirow{2}{*}{\textbf{Policy}} & \multirow{2}{*}{\textbf{Venue}} & \multicolumn{4}{c}{\textbf{Closed-loop metrics}} &
      \multicolumn{4}{c}{\textbf{Closed-loop ability scores}} \\
      \cmidrule(lr){3-6}\cmidrule(l){7-10}
      & & \textbf{DS} & \textbf{SR} & \textbf{DE} & \textbf{Comf.} & \textbf{TC} & \textbf{VRUC} & \textbf{TV} & \textbf{TI} \\
      \midrule
      \textbf{\emph{Human Driver Reference}} & -- & 100.00{\tiny$\pm$0.00} & 100.00{\tiny$\pm$0.00} & 316.52{\tiny$\pm$0.42} & 75.15{\tiny$\pm$1.89} & 100.00{\tiny$\pm$0.00} & 100.00{\tiny$\pm$0.00} & 100.00{\tiny$\pm$0.00} & 100.00{\tiny$\pm$0.00} \\
      \midrule
      \multicolumn{10}{@{}l}{\emph{Privileged reference}} \\
      PDM-Closed~\citep{Dauner2023PDM} & CoRL 2023 & 90.17{\tiny$\pm$1.00} & 81.31{\tiny$\pm$1.03} & 305.27{\tiny$\pm$8.57} & 82.49{\tiny$\pm$1.39} & 88.67{\tiny$\pm$1.53} & 72.50{\tiny$\pm$0.00} & 84.24{\tiny$\pm$1.39} & 57.78{\tiny$\pm$1.92} \\
      \midrule
      \multicolumn{10}{@{}l}{\emph{IL-based Methods}} \\
      LTF~\citep{Dauner2024NAVSIM} & NeurIPS 2024 & 53.04{\tiny$\pm$1.92} & 27.74{\tiny$\pm$1.76} & 163.71{\tiny$\pm$0.11} & \cellone{97.81{\tiny$\pm$0.15}} & 31.67{\tiny$\pm$0.58} & 25.00{\tiny$\pm$7.50} & 28.48{\tiny$\pm$1.89} & 15.56{\tiny$\pm$1.92} \\
      DiffusionDrive~\citep{Liao2025DiffusionDrive} & CVPR 2025 & 44.77{\tiny$\pm$0.75} & 20.95{\tiny$\pm$0.55} & 165.08{\tiny$\pm$1.84} & \celltwo{97.68{\tiny$\pm$0.56}} & 21.33{\tiny$\pm$1.15} & 11.67{\tiny$\pm$1.44} & 23.64{\tiny$\pm$0.00} & 22.22{\tiny$\pm$1.92} \\
      DrivoR~\citep{Kirby2026DrivoR} & CVPR 2026 & \cellthree{64.74{\tiny$\pm$1.60}} & \celltwo{50.71{\tiny$\pm$0.71}} & 188.02{\tiny$\pm$1.07} & 87.92{\tiny$\pm$1.25} & \celltwo{58.67{\tiny$\pm$2.52}} & 31.67{\tiny$\pm$5.77} & \cellthree{52.42{\tiny$\pm$2.62}} & 43.33{\tiny$\pm$0.00} \\
      RAP~\citep{Feng2026RAP} & ICLR 2026 & 47.70{\tiny$\pm$0.66} & 27.26{\tiny$\pm$0.41} & \celltwo{220.78{\tiny$\pm$3.77}} & 48.38{\tiny$\pm$0.60} & 32.33{\tiny$\pm$1.15} & 14.17{\tiny$\pm$6.29} & 28.79{\tiny$\pm$0.52} & 22.22{\tiny$\pm$3.85} \\
      SparseDriveV2~\citep{Sun2026SparseDriveV2} & ECCV 2026 & \celltwo{65.26{\tiny$\pm$1.72}} & 48.81{\tiny$\pm$3.04} & 167.99{\tiny$\pm$1.01} & 93.41{\tiny$\pm$0.43} & \cellthree{54.00{\tiny$\pm$0.00}} & \celltwo{43.33{\tiny$\pm$8.04}} & 48.18{\tiny$\pm$4.17} & 41.11{\tiny$\pm$3.85} \\
      GTRS-Dense--V2-99~\citep{Li2025GTRS} & arXiv 2025 & 64.27{\tiny$\pm$0.76} & \cellthree{49.52{\tiny$\pm$1.15}} & 163.88{\tiny$\pm$0.87} & 95.90{\tiny$\pm$0.63} & 52.00{\tiny$\pm$1.73} & \cellthree{40.00{\tiny$\pm$2.50}} & \celltwo{54.24{\tiny$\pm$0.52}} & 36.67{\tiny$\pm$0.00} \\
      ReCogDrive--2B--IL~\citep{Li2025ReCogDrive} & ICLR 2026 & 48.98{\tiny$\pm$0.95} & 28.21{\tiny$\pm$1.43} & 166.94{\tiny$\pm$3.15} & \cellthree{97.58{\tiny$\pm$0.11}} & 30.00{\tiny$\pm$1.00} & 16.67{\tiny$\pm$3.82} & 31.21{\tiny$\pm$0.52} & 26.67{\tiny$\pm$3.33} \\
      MTDrive--SFT~\citep{Li2026MTDrive} & arXiv 2026 & 48.78{\tiny$\pm$0.88} & 31.79{\tiny$\pm$1.43} & 173.77{\tiny$\pm$0.76} & 94.51{\tiny$\pm$1.25} & 30.33{\tiny$\pm$1.15} & 15.00{\tiny$\pm$6.61} & 35.76{\tiny$\pm$1.39} & \cellthree{44.44{\tiny$\pm$1.92}} \\
      \midrule
      \multicolumn{10}{@{}l}{\emph{RLFT-based Methods}} \\
      ReCogDrive--2B--RL~\citep{Li2025ReCogDrive} & ICLR 2026 & 53.64{\tiny$\pm$1.42} & 30.95{\tiny$\pm$2.06} & 173.96{\tiny$\pm$1.90} & 85.10{\tiny$\pm$0.49} & 35.33{\tiny$\pm$2.31} & 18.33{\tiny$\pm$6.29} & 32.42{\tiny$\pm$1.39} & 27.78{\tiny$\pm$5.09} \\
      MTDrive--mtGRPO~\citep{Li2026MTDrive} & arXiv 2026 & 55.86{\tiny$\pm$0.50} & 40.60{\tiny$\pm$0.90} & \cellone{277.62{\tiny$\pm$3.17}} & 93.94{\tiny$\pm$0.41} & 40.33{\tiny$\pm$2.89} & 16.67{\tiny$\pm$1.44} & 46.36{\tiny$\pm$0.91} & \cellone{52.22{\tiny$\pm$3.85}} \\
      AutoVLA~\citep{Zhou2025AutoVLA} & NeurIPS 2025 & 51.05{\tiny$\pm$0.80} & 29.52{\tiny$\pm$2.43} & 186.02{\tiny$\pm$0.58} & 65.37{\tiny$\pm$0.47} & 33.00{\tiny$\pm$2.65} & 15.00{\tiny$\pm$4.33} & 29.70{\tiny$\pm$2.92} & 36.67{\tiny$\pm$8.82} \\
      \midrule
      \multicolumn{10}{@{}l}{\emph{Passive Demonstration Perturbation Methods}} \\
      DiffusionDrive~{\scriptsize(SimScale)}~\citep{Tian2026SimScale} & CVPR 2026 & 52.19{\tiny$\pm$2.05} & 35.24{\tiny$\pm$2.38} & 189.66{\tiny$\pm$4.28} & 96.92{\tiny$\pm$0.34} & 38.33{\tiny$\pm$1.15} & 18.33{\tiny$\pm$7.22} & 36.67{\tiny$\pm$5.01} & 42.22{\tiny$\pm$3.85} \\
      DrivoR~{\scriptsize(SimScale, 134k)}~\citep{Tian2026SimScale} & CVPR 2026 & 57.66{\tiny$\pm$1.32} & 42.50{\tiny$\pm$0.36} & 176.73{\tiny$\pm$0.92} & 85.88{\tiny$\pm$0.39} & 44.00{\tiny$\pm$1.00} & 17.50{\tiny$\pm$2.50} & 48.18{\tiny$\pm$2.41} & \celltwo{50.00{\tiny$\pm$0.00}} \\
      LTF~{\scriptsize(SimScale)}~\citep{Tian2026SimScale} & CVPR 2026 & 52.65{\tiny$\pm$0.88} & 34.05{\tiny$\pm$0.21} & \cellthree{192.09{\tiny$\pm$0.85}} & 97.41{\tiny$\pm$0.48} & 35.00{\tiny$\pm$1.73} & 17.50{\tiny$\pm$2.50} & 38.48{\tiny$\pm$0.52} & 36.67{\tiny$\pm$0.00} \\
      GTRS-Dense--V2-99~{\scriptsize(SimScale)}~\citep{Tian2026SimScale} & CVPR 2026 & \cellone{73.54{\tiny$\pm$0.84}} & \cellone{61.79{\tiny$\pm$1.56}} & 141.50{\tiny$\pm$1.30} & 87.69{\tiny$\pm$1.66} & \cellone{69.33{\tiny$\pm$0.58}} & \cellone{50.00{\tiny$\pm$2.50}} & \cellone{64.24{\tiny$\pm$1.89}} & 43.33{\tiny$\pm$3.33} \\
      DiffusionDrive~{\scriptsize(BeyondDrive)}~\citep{Wang2026BeyondDrive} & ECCV 2026 & 46.61{\tiny$\pm$0.07} & 21.79{\tiny$\pm$0.94} & 172.79{\tiny$\pm$1.26} & 97.24{\tiny$\pm$0.19} & 22.67{\tiny$\pm$1.53} & 10.00{\tiny$\pm$0.00} & 25.15{\tiny$\pm$1.39} & 22.22{\tiny$\pm$1.92} \\
      \midrule
      \multicolumn{10}{@{}l}{\emph{World-Model-based Methods}} \\
      SimWAM~\citep{Zhao2026SimWAM} & arXiv 2026 & 57.53{\tiny$\pm$2.72} & 37.50{\tiny$\pm$1.56} & 160.54{\tiny$\pm$1.64} & 97.19{\tiny$\pm$0.27} & 41.33{\tiny$\pm$0.58} & 26.67{\tiny$\pm$7.22} & 40.61{\tiny$\pm$0.52} & 27.78{\tiny$\pm$5.09} \\
      \bottomrule
      \end{tabular}
      }
  \end{table*}

The leaderboard reveals two important insights. First, strong OL performance does not ensure CL safety. Despite approaching human-level performance on the NAVSIM leaderboard (Figure~\ref{fig:navsim_navsafe_gap}), the evaluated policies exhibit a pronounced CL safety gap: even the strongest learned policy trails the privileged PDM-Closed reference by $19.5$ percentage points in SR. Second, safety performance varies substantially across event types, and no learning-based policy leads all four categories. The same training intervention can affect policies differently: SimScale raises DiffusionDrive's SR but lowers DrivoR's (Table~\ref{tab:main}). A single OL or aggregate CL score therefore cannot reveal how a policy behaves across the states and safety events encountered during interaction.

\subsection{Failure Investigations}
\label{sec:failure_investigations}

We further conduct case studies to investigate failure modes across representative E2E policy paradigms. We analyze one diagnostic case from each of the four \navsafe{} categories, designed to probe distinct safety capabilities. This analysis localizes unsafe CL outcomes to specific breakdowns in \emph{planning, scoring, prediction, and reasoning}. Case details are in Appendix~\ref{app:failure_case_setting}.

\begin{figure*}[tbh]
  \centering
  \includegraphics[width=\textwidth]{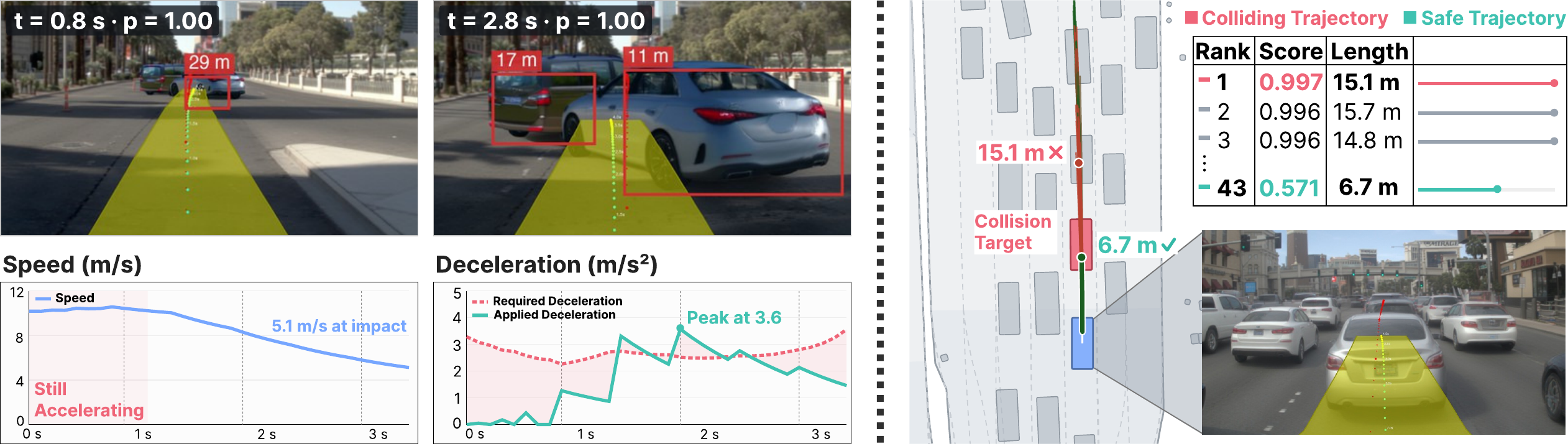}
  \caption{\textbf{Left: Planning Failure.}
    LTF detects crossing traffic at $29\,\mathrm{m}$, but continues accelerating and brakes too late to avoid a collision.
    \textbf{Right: Scoring Failure.}
    DrivoR executes its top-ranked trajectory and collides with the lead vehicle, while a safe stopping trajectory is ranked 43rd.}
  \label{fig:failure_analysis_ltf_drivor}
\end{figure*}

\textbf{Planning Failure: Causal confusion in hazard-conditioned planning.}
The I-3 General Incident event in the Traffic Incidents (TI) category requires a prompt response to an unexpected crossing conflict. Although LTF consistently detects the crossing vehicles, its longitudinal plan resembles nominal car following and delays decisive braking until the collision becomes unavoidable (Figure~\ref{fig:failure_analysis_ltf_drivor}, left). This behavior is consistent with causal confusion in offline behavior cloning: the policy perceives the hazard, yet its plan remains governed by correlations associated with routine car following rather than by the detected crossing conflict.

\textbf{Scoring Failure: Trajectory scoring gap prevents safe planning.}
The V-1 Red-Light/Stop-Sign event in the Traffic Violations (TV) category requires the policy to stop at the applicable traffic control. DrivoR's candidate set contains a collision-free stopping trajectory, yet the learned scorer assigns a higher value to the collision-bound proposal (Figure~\ref{fig:failure_analysis_ltf_drivor}, right). The available safe proposal localizes the observed failure to trajectory ranking rather than candidate generation. DrivoR learns its scoring function from trajectory-level oracle labels evaluated at logged states. The observed misranking exposes an OL--CL generalization gap in which the learned safety ordering does not transfer to the policy-induced state encountered during rollout. 

\begin{figure}[tbh]
    \centering
    \includegraphics[width=\linewidth]{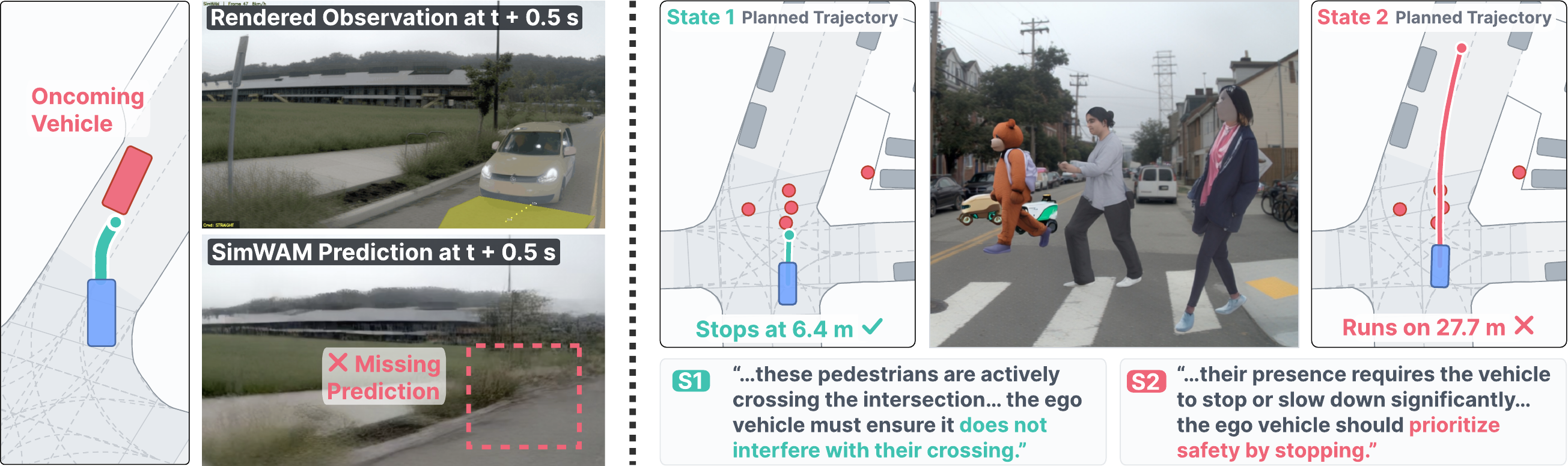}
    \caption{\textbf{Left: Prediction Failure.} SimWAM omits the
    oncoming vehicle visible in the rendered observation; the policy plans
    through the space it occupies and collides.
    \textbf{Right: Reasoning Failure.} At the same instant, two policy-induced states reached by different rollouts both identify the crossing pedestrians and call for a stop, but only State~1 plans to stop; State~2 accelerates and collides.}
    \label{fig:failure_analysis_vla_wm}
\end{figure}

\textbf{Prediction Failure: Latent prediction errors can undermine planning in world-model-based policies.}
The C-7 Head-On event in the Traffic Crashes (TC) category requires the policy to maintain sufficient clearance from oncoming traffic in a constrained corridor. For this diagnostic, we re-enable and decode a future frame with SimWAM's retained video expert from the same CL observation. SimWAM omits the oncoming vehicle from its future prediction and plans through the region it occupies (Figure~\ref{fig:failure_analysis_vla_wm}, left). This coupled failure exposes a breakdown in action-relevant actor persistence, as the model drops a safety-critical actor from its predicted world state. Latent prediction errors can undermine world-model-based planning by removing safety-critical constraints from the representation used for trajectory generation under CL execution.

\textbf{Reasoning Failure: Reasoning--action alignment is unstable across policy-induced states for VLA policies.}
The R-1 Pedestrian-Involved event in the Vulnerable Road User Crashes (VRUC) category requires the policy to stop or yield to pedestrians crossing the ego path. At the same instant, in two rollouts that differ only in the evaluation seed, AutoVLA correctly calls for stopping in two policy-induced states, but only one decoded trajectory realizes that decision (Figure~\ref{fig:failure_analysis_vla_wm}, right). The matched comparison isolates a state-conditioned failure of reasoning--action grounding, whereby the reasoning remains invariant while the resulting plans diverge across policy-induced states.

\subsection{Understanding the Limits of Passive Demonstration Perturbation and RLFT in Closed-Loop}
\label{sec:method_diagnosis}

Passive demonstration perturbation and RLFT are two widely adopted remedies that go beyond imitation learning to improve CL performance~\citep{Karkus2025BeyondBC}. Perturbation enhances IL by further training the policy on states constructed independently of its own rollouts. We show that this approach helps mainly when the policy's CL rollouts remain close to these perturbed states. RLFT for E2E driving, on the other hand, typically optimizes an OL proxy reward against a frozen logged future. As a result, OL reward hacking, such as trading safety margin for progress, can carry over to CL execution and compound over repeated replanning steps.

\begin{figure}[t]
  \centering
  \includegraphics[width=\linewidth]{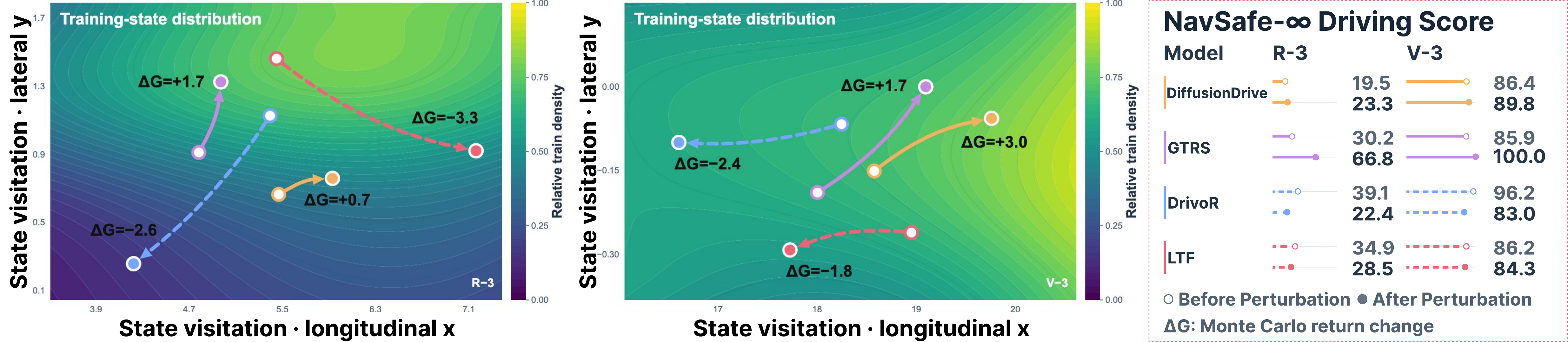}
  \caption{\textbf{Effect of perturbation on CL rollouts.} State shifts and Monte Carlo return changes show that perturbation helps when CL rollout states remain covered and hurts otherwise (Appendix~\ref{app:perturbation_setting}).}
    \label{fig:augmentation_support}
\end{figure}

\textbf{Perturbation effectiveness hinges on CL rollout distribution coverage.}
Such methods~\citep{Tian2026SimScale} extend the logged demonstrations $\mathcal{D}^{\text{log}}$ with $\mathcal{D}^{\text{aug}}$, pairing observations of perturbed ego states $s^{\text{aug}}$ with verified expert recovery trajectories $a^{\text{aug}}$:
\begin{equation}
\min_{\theta}\;
\mathbb{E}_{(s,a)\sim\mathcal{D}^{\text{log}}\cup\,{\mathcal{D}^{\text{aug}}}}\!\left[\ell\big(a,\pi_\theta(\cdot\mid s)\big)\right].
\label{eq:passive_perturbation}
\end{equation}
This improves CL performance when rollouts remain near the perturbation support, but this depends on the learned trajectory proposal, which the objective does not directly constrain.
In practice, the outcome varies across policies (Figure~\ref{fig:augmentation_support}). 
For anchor-based methods such as DiffusionDrive~\citep{Liao2025DiffusionDrive} and GTRS~\citep{Li2025GTRS}, trajectory candidates remain close to expert trajectories and are largely unaffected by training on perturbed states. Training mainly improves the scorer, keeping selected rollouts closer to the perturbed state distribution. For query-based DrivoR and LTF, training on $\mathcal{D}^{\text{aug}}$ improves OL scores on expert states, yet CL visitation still drifts away from the support of $\mathcal{D}^{\text{aug}}$. This limitation arises because $\mathcal{D}^{\text{aug}}$ is constructed independently of the policy and thus does not capture the policy visitation distribution, which only emerges in CL execution.

\begin{figure}[t]
  \centering
  \includegraphics[width=\linewidth]{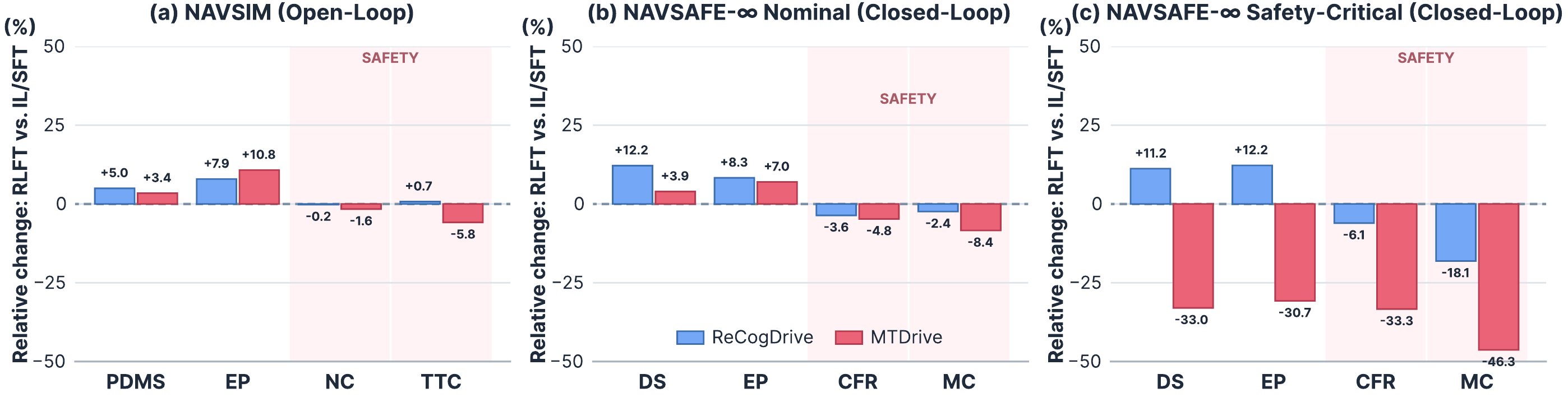}
  \caption{\textbf{RLFT reward exploitation from OL to CL.} (a) OL RLFT reward hacking trades safety for progress. (b) The trade-off persists in nominal CL planning. (c) In safety-critical scenarios, the resulting collisions terminate rollouts early and amplify the degradation in CL performance. EP: Ego Progress in (a), Event Progress in (b) and (c); NC: no at-fault collision; TTC: time to collision; DS: driving score; CFR: collision-free rate; MC: minimum clearance.}
  \label{fig:rlft_ol_cl_safety_gap}
\end{figure}

\textbf{OL reward hacking transfers and compounds in CL execution.} Most existing RLFT methods~\citep{Li2025ReCogDrive,Li2026MTDrive} optimize an OL objective on states visited by the expert, scoring each sampled trajectory once with a proxy reward $R'$, such as PDMS~\citep{Dauner2024NAVSIM}, against a frozen logged future:
\begin{equation}
\max_{\theta}\;
\mathbb{E}_{s^{\text{log}}\sim\mathcal{D}^{\text{log}},\;a\sim\pi_\theta(\cdot\mid s^{\text{log}})}\!\left[R'\big(s^{\text{log}},a\big)\right].
\label{eq:ol_rlft}
\end{equation}
Because $R'$ rewards progress directly but constrains safety margin only weakly, the policy can raise $R'$ by trading margin for progress. This exploitation carries over to CL, raising the per-step risk of a safety-critical error; in safety-critical scenarios, such errors compound over the rollout and lower the CL return. The two RLFT policies we evaluate illustrate both regimes (Figure~\ref{fig:rlft_ol_cl_safety_gap}; Appendix~\ref{app:diagnostic_setting}). RLFT as applied in ReCogDrive leaves the safety-related sub-scores largely intact, so its OL gains transfer to CL. RLFT as applied in MTDrive instead raises PDMS mainly through higher progress at the cost of lower safety scores, causing a sharp drop in CL DS under safety-critical scenarios.

\section{\navsafe{} Toolbox}
\label{sec:toolbox}

To facilitate further research on CL policy behavior, we provide an extensible \navsafe{} toolbox that enables researchers to \textbf{(1)} build photorealistic CL environments from additional real-world driving logs, \textbf{(2)} create controlled tests for user-defined safety events, and \textbf{(3)} examine how new policies behave under off-nominal states and evolving traffic interactions.

\textbf{(1) Photorealistic environment construction.} The environment-construction tool converts real-world driving logs~\citep{Caesar2021nuPlan,Sun2020Waymo} into interactive simulation environments. It maps calibrated sensor streams, ego poses, actor tracks, traffic-light states, and map context into a unified simulation representation, and reconstructs the corresponding visual scene as a 3DGS environment~\citep{NVIDIA2026NuRec,Huang2026InstantNuRec}.

\textbf{(2) Customizable scenarios.}
The event-design tool provides an asset zoo of vehicles, vulnerable road users, traffic-control devices, and roadside hazards, plus 3DGS assets generated from a reference image~\citep{Cao2026AssetHarvester}. Assets are inserted into a reconstructed scene with user-specified size, pose, and a static, scripted, or reactive behavior, enabling researchers to create events absent from the log and controlled variants that change one factor at a time.

\textbf{(3) Closed-loop policy diagnosis.}
The diagnosis tool records policy-induced ego and environment states in a rollout archive, from which controlled counterfactual evaluations can be initialized. Ego-state interventions alter the vehicle's pose or motion state, whereas environment-state interventions modify the behavior of surrounding actors, such as through lead-vehicle braking or cut-in maneuvers. These controlled replays reveal whether policy improvements persist as CL rollouts move beyond the states represented in logged or perturbed demonstrations.

\section{Limitations}
\label{sec:limitations}

\paragraph{Sim2Real Interpretation.} Although photorealistic rendering narrows the visual-domain gap, the simulator cannot fully capture sensor noise, reconstruction artifacts, contact dynamics, or human driving behavior. Consequently, failures may arise from simulator mismatch, while successful rollouts may benefit from simplified agent responses. The results of \navsafe{} should therefore be interpreted as controlled evidence within the evaluated simulation setting rather than as direct estimates of real-world crash risk. 

\paragraph{Scenario Evaluation Coverage.} Our benchmark primarily focuses on dense urban driving scenarios~\citep{Dauner2024NAVSIM, Caesar2021nuPlan}, with highway, rural, adverse-weather, and nighttime conditions comparatively underrepresented. Expanding \navsafe{} to a broader range of driving environments and conditions is an important direction for future evaluation.

\section{Conclusion}
\label{sec:conclusion}

E2E driving safety must be assessed through policy-induced interactions.
\navsafe{} enables it through photorealistic and event-level CL evaluation, revealing specific behavioral failures that non-reactive evaluations can obscure. Our results show that strong OL performance does not reliably translate into CL safety.
The analysis attributes these failures to causal confusion, scorer bottlenecks, latent dynamic discrepancy, and reasoning-action misalignment. 
Common post-IL remedies do not reliably resolve these failures. Passive demonstration perturbation helps mainly when CL rollouts remain near the perturbed support, while reward hacking on expert states with RLFT carries over to CL and compounds through feedback. 
\navsafe{} will be actively maintained to facilitate future research on CL driving.

\subsection*{AI Use Statement}

In this work, we used generative AI tools for manuscript editing and assistance with software development and debugging. We did not use generative AI tools to formulate scientific claims, design the evaluation methodology, or interpret experimental results. We manually reviewed and tested AI-assisted code for correctness, and reviewed all AI-assisted work. We take responsibility for the final content of this work, including text, claims, or artifacts produced with the aid of generative AI.

\subsection*{Ethics Statement}

This work evaluates autonomous-driving policies entirely in simulation and does not deploy vehicles on public roads or collect new data from road users. The benchmark uses publicly released driving logs, third-party model checkpoints, and NVIDIA NRE/NuRec artifacts under their respective licenses and access conditions; photographs used to construct animal and prohibitory-sign assets were captured by the authors. The Human Driver Reference was produced under an IRB-approved protocol by three participants, each of whom completed all 280 scenarios under one assigned seed using a game controller. All participants provided written informed consent, and separate consent was obtained for publication of the evaluation-setup photograph. No sensitive personal data are reported. The human results establish that the evaluated scenario--seed cells are solvable under the specified interface; they are not intended to characterize population-level human driving performance or between-driver variability. Finally, \navsafe{} is intended for controlled research evaluation: simulation scores should not be interpreted as certification of deployment readiness or as estimates of real-world crash risk.

\subsection*{Reproducibility Statement}

Appendices~\ref{app:benchmark_details}--\ref{app:reproducibility} document the benchmark taxonomy and scenario provenance, asset construction, scene reconstruction and rendering, evaluation protocol and metrics, evaluated checkpoints and policy adapters, diagnostic experiments, toolbox implementation, and runtime requirements. Every policy is evaluated on the same 280 scenarios under fixed seeds 0, 1, and 1024, with means and sample standard deviations reported across seeds. Each evaluation records its resolved configuration, checkpoint identifier, requested trajectory, state trace, infractions, and terminal reason, allowing metrics to be recomputed from saved traces. Upon publication, we will release the simulator code, benchmark recipes and manifests, policy adapters, evaluation configurations, and scripts needed to reproduce the reported tables. Third-party datasets, model weights, and NVIDIA NRE/NuRec components remain subject to their original licenses and access requirements.

\ificlrfinal
\fi

\clearpage
\bibliography{iclr2027_conference}
\bibliographystyle{iclr2027_conference}

\newpage
\appendix

\section*{Appendix}
\label{app:overview}

\noindent The appendix is organized as follows.
\begin{itemize}\setlength{\itemsep}{0pt}
  \item Appendix~\ref{app:benchmark_details}: Scenario taxonomy, coverage, and scenario gallery (Section~\ref{sec:event_design}).
  \item Appendix~\ref{app:assets}: Asset library used for scenario interventions.
  \item Appendix~\ref{app:environment_details}: Dataset source, photorealistic reconstruction, and rendering fidelity (Section~\ref{sec:rendering_fidelity}).
  \item  Appendix~\ref{app:evaluation_protocol}: Closed-loop protocol, event contracts, metrics, and benchmark quality controls
  (Section~\ref{sec:protocol}).
  \item Appendix~\ref{app:experimental_details}: Evaluated policies and the human driver reference (Table~\ref{tab:main}).
  \item  
Appendix~\ref{app:additional_results}: OL--CL comparison and full leaderboard, case-selection protocol for the failure investigations, and diagnostic settings (Sections~\ref{sec:leaderboard}, \ref{sec:failure_investigations}, and~\ref{sec:method_diagnosis}).
  \item Appendix~\ref{app:toolbox_details}: Toolbox implementation (Section~\ref{sec:toolbox}).
  \item Appendix~\ref{app:reproducibility}: Runtime and resource requirements.
\end{itemize}

\section{\navsafe{} Scenario Taxonomy and Gallery}
\label{app:benchmark_details}

This appendix supplements Section~\ref{sec:event_design} with detailed operational criteria, nuPlan screening labels, scenario-construction routes, and per-event scenario counts. Together with Figure~\ref{fig:scenario_demo}, the additional examples provide one rendered scenario for each of the 28 event types.

\subsection{Taxonomy Definitions and Scenario Coverage}
\label{app:taxonomy_coverage}

Each of the 280 scenarios belongs to exactly one of 28 event types adapted from RoadSafe365~\citep{RoadSafe365} (Table~\ref{tab:taxonomy_coverage}). An event type is defined by its operational inclusion criterion, which specifies the event that must occur within the evaluated window. nuPlan scenario labels~\citep{Caesar2021nuPlan} are used only to retrieve candidate windows and do not determine membership.

Scenarios are obtained through one of two routes. For 19 \emph{direct} event types, nuPlan scenario labels expose the defining event, and scenarios are retrieved from the logs. For the remaining nine \emph{recipe} event types (C-7, C-10, V-8, V-10, V-11, R-2, R-3, R-4, and I-3), no label retrieves the event reliably, so every scenario is constructed from a versioned recipe (Appendix~\ref{app:toolbox_authoring}). A recipe either retains the host log unchanged or applies an \emph{intervention}, namely a controlled actor, asset, or traffic-control edit to a compatible log.

Each event type contains ten scenarios. The 19 direct event types contribute 190 natural scenarios. The nine recipe event types contribute 90 scenarios, of which 62 are intervened and 28 replay the host log unchanged (all 20 scenarios of C-10 and V-11, and the natural scenarios of C-7, V-10, and R-2). The benchmark therefore contains 218 natural and 62 intervened scenarios. The benchmark manifest records the source log token, event type, provenance, recipe identifier, and event contract of every scenario.

\begingroup
\scriptsize
\setlength{\LTleft}{0pt}
\setlength{\LTright}{0pt}
\setlength{\tabcolsep}{2.5pt}
\renewcommand{\arraystretch}{1.10}
\setlength{\emergencystretch}{1em}

\newcommand{\nplabel}[1]{\nolinkurl{#1}}

\begin{longtable}{@{}%
>{\raggedright\arraybackslash}p{0.11\textwidth}%
>{\raggedright\arraybackslash}p{0.235\textwidth}%
>{\raggedright\arraybackslash}p{0.37\textwidth}%
>{\raggedright\arraybackslash}p{0.16\textwidth}%
>{\centering\arraybackslash}p{0.05\textwidth}@{}}

\caption{Operational taxonomy, nuPlan screening labels, and coverage route.
\emph{Direct} event types are taken from unmodified nuPlan logs (screened by
scenario type where one applies);
\emph{Recipe} event types are constructed from versioned recipes.}
\label{tab:taxonomy_coverage}\\

\toprule
\textbf{Event type} &
\textbf{Operational inclusion criterion} &
\textbf{nuPlan screening labels} &
\textbf{Coverage route} &
\shortstack{\textbf{$N$}\\\textbf{(nat./int.)}}\\
\midrule
\endfirsthead

\toprule
\textbf{Event type} &
\textbf{Operational inclusion criterion} &
\textbf{nuPlan screening labels} &
\textbf{Coverage route} &
\shortstack{\textbf{$N$}\\\textbf{(nat./int.)}}\\
\midrule
\endhead

\midrule
\multicolumn{5}{r@{}}{\textit{Continued on next page}}\\
\endfoot

\bottomrule
\endlastfoot

\addlinespace[2pt]
\multicolumn{5}{@{}l}{\textit{Traffic Crashes}}\\*[-1pt]

C-1 Rear-End &
Same-lane longitudinal conflict in which inadequate headway or braking can
produce front-to-rear contact. &
\nplabel{stopping_with_lead};
\nplabel{behind_long_vehicle};
\nplabel{near_long_vehicle};
\nplabel{stopping_at_traffic_light_with_lead};
\nplabel{stopping_at_stop_sign_with_lead};
\nplabel{stationary_at_traffic_light_with_lead};
\nplabel{accelerating_at_traffic_light_with_lead}. &
Direct. &
10/0\\

C-2 Angle/T-Bone &
Crossing paths create a lateral-impact conflict at an intersection or access
point. &
\nplabel{crossed_by_vehicle};
\nplabel{starting_protected_cross_turn}. &
Direct. &
10/0\\

C-3 Sideswipe &
Overlapping lateral motion between adjacent vehicles creates same-direction
side-contact risk. &
\nplabel{changing_lane_with_lead};
\nplabel{changing_lane_with_trail};
\nplabel{near_high_speed_vehicle}. &
Direct. &
10/0\\

C-4 Multi-Vehicle &
At least four interacting vehicles permit a primary conflict to propagate
into secondary braking or collision risk. &
\nplabel{near_multiple_vehicles}. &
Direct. &
10/0\\

C-5 Intersection &
The safety-critical conflict and required maneuver occur within an
intersection footprint. &
\nplabel{traversing_intersection};
\nplabel{on_intersection};
\nplabel{starting_left_turn};
\nplabel{starting_right_turn};
\nplabel{starting_low_speed_turn};
\nplabel{starting_high_speed_turn};
\nplabel{starting_protected_noncross_turn}. &
Direct. &
10/0\\

C-6 Roadway Departure &
Ego must re-enter the through lane from a curbside or pickup/drop-off position. &
\nplabel{traversing_pickup_dropoff};
\nplabel{on_pickup_dropoff}. &
Direct. &
10/0\\

C-7 Head-On &
Opposing traffic competes for insufficient longitudinal or lateral clearance
in a shared corridor. &
\nplabel{traversing_narrow_lane}. &
Recipe: natural log plus targeted oncoming actor. &
2/8\\

C-8 Pull-Over &
Ego must leave the through lane into a curbside or parking position, where the low-speed lateral transition creates contact risk with objects or road users near the boundary. &
\nplabel{on_carpark}. &
Direct. &
10/0\\

C-9 Single-Vehicle &
No other traffic participant is present in the ego's conflict region, so any safety event arises from the ego's own control, the surface, or a fixed object. &
None. &
Direct. &
10/0\\

C-10 Wrong-Way Crash &
Ego enters a lane assigned to the opposite direction of travel, so oncoming traffic produces a head-on conflict. &
None. &
Recipe: natural road geometry. &
10/0\\

\addlinespace[3pt]
\multicolumn{5}{@{}l}{\textit{Traffic Violations}}\\*[-1pt]

V-1 Red-Light/Stop-Sign &
A traffic light or stop sign governs ego entry at a control line. &
\nplabel{traversing_traffic_light_intersection};
\nplabel{on_traffic_light_intersection};
\nplabel{on_stopline_traffic_light};
\nplabel{starting_straight_traffic_light_intersection_traversal};
\nplabel{stopping_at_traffic_light_without_lead};
\nplabel{stationary_at_traffic_light_without_lead};
\nplabel{accelerating_at_traffic_light};
\nplabel{accelerating_at_traffic_light_without_lead};
\nplabel{on_stopline_stop_sign};
\nplabel{starting_straight_stop_sign_intersection_traversal};
\nplabel{stopping_at_stop_sign_without_lead};
\nplabel{stopping_at_stop_sign_no_crosswalk};
\nplabel{accelerating_at_stop_sign};
\nplabel{accelerating_at_stop_sign_no_crosswalk}. &
Direct. &
10/0\\

V-2 Failure to Yield &
Ego must reject or accept a gap under an explicit right-of-way obligation. &
\nplabel{starting_unprotected_cross_turn};
\nplabel{starting_unprotected_noncross_turn};
\nplabel{on_all_way_stop_intersection}. &
Direct. &
10/0\\

V-3 Unsafe Lane Change &
A lane transfer or pass is unsafe because of clearance, boundary, signaling,
or maneuver-completion constraints. &
\nplabel{changing_lane};
\nplabel{changing_lane_to_left};
\nplabel{changing_lane_to_right}. &
Direct. &
10/0\\

V-4 Reckless/Aggressive &
Ego action exceeds a preregistered kinematic or interaction-risk envelope
without necessity. &
\nplabel{high_lateral_acceleration};
\nplabel{high_magnitude_jerk}. &
Direct. &
10/0\\

V-5 Speeding &
Ego speed violates the applicable posted or scenario-specific limit. &
\nplabel{high_magnitude_speed};
\nplabel{medium_magnitude_speed};
\nplabel{low_magnitude_speed}. &
Direct. &
10/0\\

V-6 Tailgating &
Ego follows a lead below the declared time-headway or stopping-distance
threshold. &
\nplabel{following_lane_with_lead};
\nplabel{following_lane_with_slow_lead}. &
Direct. &
10/0\\

V-7 Lane Maintenance &
Uncommanded lateral drift crosses or persistently approaches the lane or
drivable-corridor boundary. &
\nplabel{following_lane_without_lead}. &
Direct. &
10/0\\

V-8 Illegal Turn &
A turn is geometrically feasible but prohibited by the applicable
control or lane rule. &
\nplabel{starting_u_turn}. &
Recipe: prohibitory-sign intervention. &
0/10\\

V-9 Blocking Intersection &
Ego enters without downstream clearance or remains stopped in a conflict
area without necessity. &
\nplabel{stationary};
\nplabel{stationary_in_traffic}. &
Direct. &
10/0\\

V-10 Unsafe Merge &
Ego enters a traffic stream without the declared longitudinal and lateral
safety margins. &
None. &
Recipe: natural log plus targeted merge actor. &
3/7\\

V-11 Wrong-Way Driving &
Ego traverses a lane against its lawful direction even without a collision. &
None. &
Recipe: natural road geometry. &
10/0\\

\addlinespace[3pt]
\multicolumn{5}{@{}l}{\textit{Vulnerable Road User Crashes}}\\*[-1pt]

R-1 Pedestrian-Involved &
A pedestrian occupies, approaches, or is occluded near the ego's potential
path and requires yielding or clearance. &
\nplabel{waiting_for_pedestrian_to_cross};
\nplabel{near_pedestrian_on_crosswalk};
\nplabel{near_pedestrian_on_crosswalk_with_ego};
\nplabel{near_multiple_pedestrians};
\nplabel{behind_pedestrian_on_driveable};
\nplabel{behind_pedestrian_on_pickup_dropoff};
\nplabel{near_pedestrian_at_pickup_dropoff};
\nplabel{traversing_crosswalk};
\nplabel{on_stopline_crosswalk};
\nplabel{stopping_at_crosswalk};
\nplabel{stationary_at_crosswalk};
\nplabel{accelerating_at_crosswalk}. &
Direct. &
10/0\\

R-2 Bicycle-Involved &
A cyclist is followed, crossed, passed, or encountered in a shared conflict
region. &
\nplabel{behind_bike};
\nplabel{crossed_by_bike};
\nplabel{near_multiple_bikes}. &
Recipe: natural log plus targeted cyclist actor. &
3/7\\

R-3 Micromobility-Involved &
A micromobility device or other atypical road user, such as a delivery robot, enters shared space.&
None. &
Recipe: targeted micromobility actor. &
0/10\\

R-4 Animal-Involved &
An animal enters or approaches the drivable corridor without
traffic-rule-compliant behavior. &
None. &
Recipe: targeted animal actor. &
0/10\\

\addlinespace[3pt]
\multicolumn{5}{@{}l}{\textit{Traffic Incidents}}\\*[-1pt]

I-1 Obstruction/Hazard &
An unmapped static object obstructs the nominal path and requires stopping
or a bounded avoidance maneuver. &
\nplabel{near_trafficcone_on_driveable};
\nplabel{near_barrier_on_driveable}. &
Direct. &
10/0\\

I-2 Work Zone &
Temporary signs, barriers, or workers alter the valid corridor relative to
the base map. &
\nplabel{near_construction_zone_sign}. &
Direct. &
10/0\\

I-3 General Incident &
A preregistered safety event not represented by another event type requires
a specific fallback response. &
None. &
Recipe: targeted incident. &
0/10\\

\end{longtable}
\endgroup

\subsection{Canonical Scenario for Each Event Type}
\label{app:scenario_gallery}

Figure~\ref{fig:all_event_types} complements the 12 event types
visualized in Figure~\ref{fig:scenario_demo} by showing one scenario
for each of the remaining 16 event types. Together, the two figures
cover all 28 taxonomy leaves. The gallery highlights the diversity of
road layouts, traffic participants, and traffic controls, while
Table~\ref{tab:taxonomy_coverage} specifies the operational inclusion criterion, nuPlan screening labels, coverage route, and natural/intervened scenario counts for each event type.

\begin{figure}[!htbp]
    \centering
    \includegraphics[width=0.94\textwidth]
    {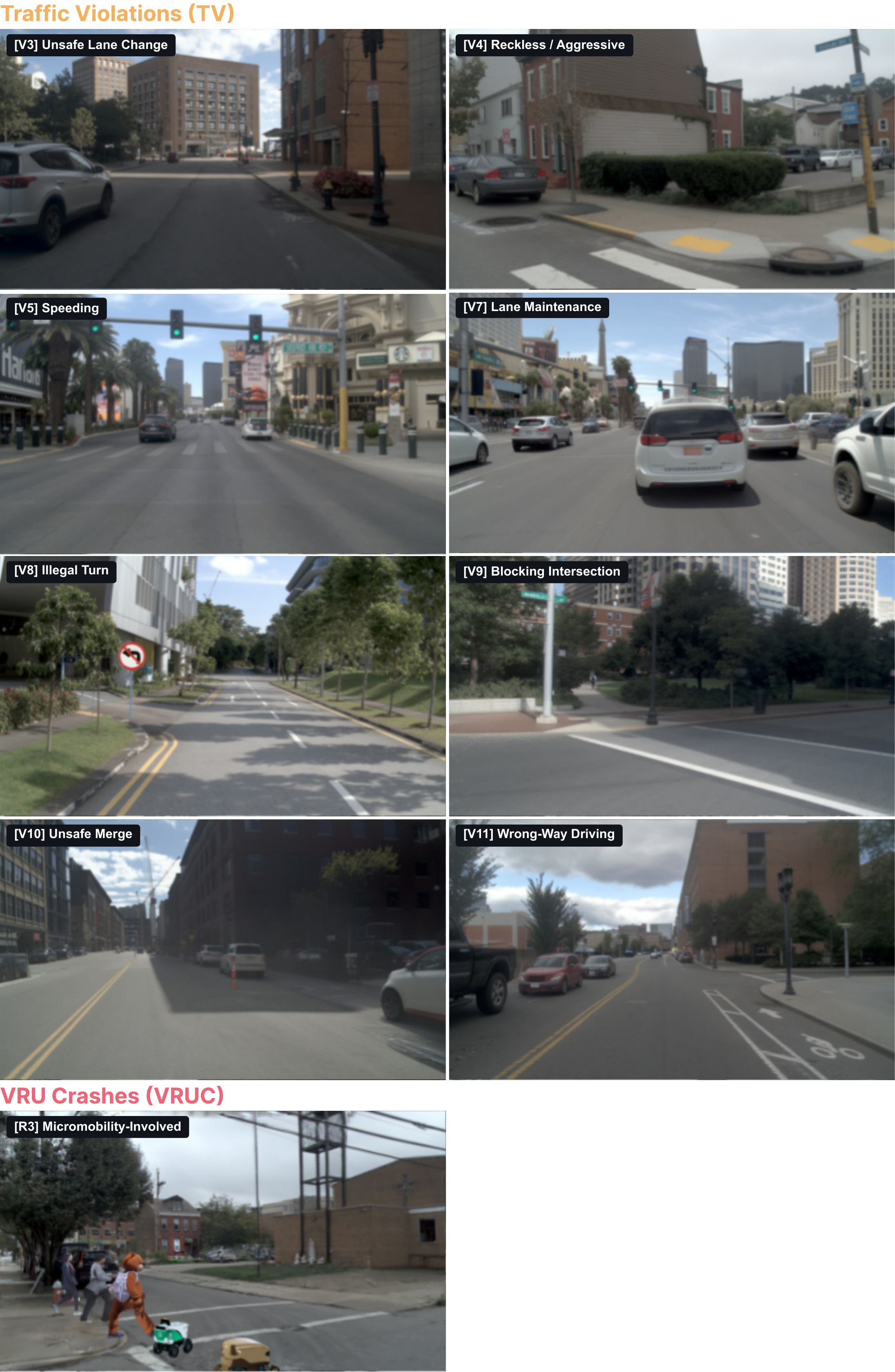}
    \caption{\textbf{Additional \navsafe{} scenario examples.}
    Complementing the 12 event types shown in
    Figure~\ref{fig:scenario_demo}, this figure presents one scenario
    for each of the remaining 16 event types; together, the two figures
    cover all 28 taxonomy leaves. Each panel shows a rendered
    front-camera observation at event onset under seed 0 using a common
    crop. The displayed scenarios were fixed before rendering and were
    not selected based on rendering quality or policy outcome.}
    \label{fig:all_event_types}
\end{figure}

\begin{figure}[t]
    \ContinuedFloat
    \centering
    \includegraphics[width=0.94\textwidth]{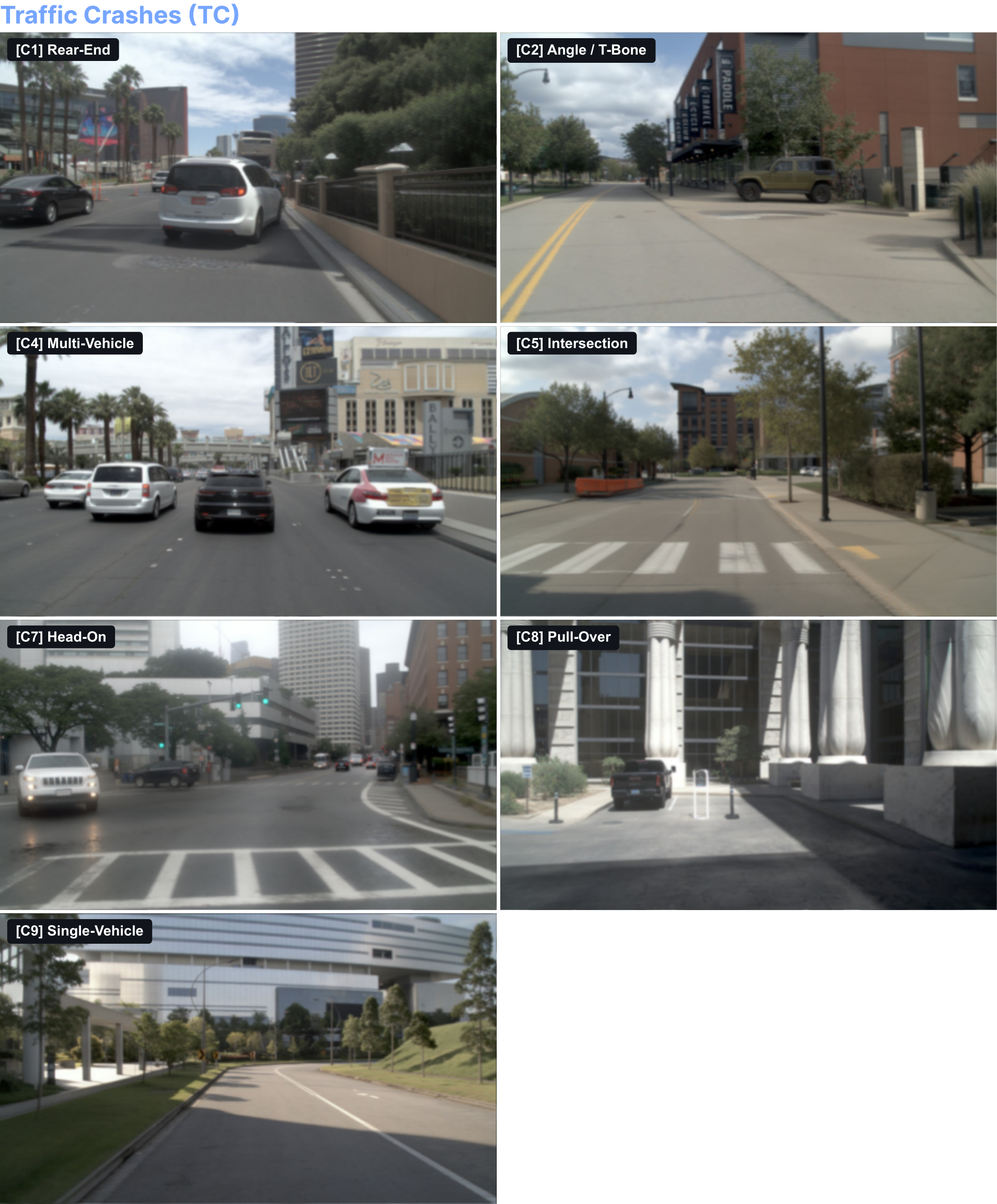}
    \caption[]{\textbf{Additional \navsafe{} scenario examples
    (continued).}}
\end{figure}

\section{\navsafe{} Asset Library}
\label{app:assets}

Intervened scenarios insert actors and objects that are absent from the host log. This appendix describes how these assets are constructed and lists the asset library. Dynamic actors follow a frozen pose track or a declared controller, and static objects are placed at a fixed pose (Appendix~\ref{app:toolbox_authoring}).

\subsection{Asset Construction}
\label{app:asset_construction}

All assets are 3DGS models generated with Asset Harvester~\citep{Cao2026AssetHarvester}. Given segmented images of an object, Asset Harvester synthesizes 16 consistent views with a multi-view diffusion model (SparseViewDiT) and lifts them to 3D Gaussians (TokenGS). Because unobserved views are generated, an asset approximates rather than reproduces the geometry and texture of its source object. The assets differ only in the source of the input images.

\textbf{Vehicles from the source logs.} We build a pipeline on Asset Harvester that converts vehicles in the nuPlan logs. Each scenario is processed as its four 5\,s clips. In each clip, vehicle tracks are projected into the calibrated cameras using their 3D bounding boxes; observations that are invalid, too small, or heavily occluded are discarded, and the remaining ones are cropped and segmented. We retain cars, trucks, buses, and trailers with at least two valid observations and a net displacement of at least 2\,m over the 20\,s scenario, and keep at most ten per scenario, ranked by their closest observation distance. Parked vehicles are excluded because the scene reconstruction already observes them from multiple viewpoints. A track that appears in several clips is converted once, from the clip in which it is observed at the closest range. Each generated asset is rotated to the NuRec coordinate convention, and assets that are empty or whose dimension ratios deviate from the source bounding box by more than 35\% are discarded. A retained asset replaces the reconstructed appearance of its source track so that the vehicle can be rendered from viewpoints not observed in the log. The asset is scaled to the bounding box of the track, and the identity and motion of the track are unchanged.

\textbf{Standalone assets.} Pedestrians, cyclists, delivery robots, and additional vehicles are converted from single nuPlan camera frames, and animals and prohibitory signs from photographs taken by the authors. Each image is converted with the single-image mode of Asset Harvester, which segments the object with its built-in segmentation model and estimates the camera viewpoint before multi-view generation and lifting. Unlike in-log vehicles, these assets are not bound to a track and can be inserted into any scenario.

\textbf{Gait animation.} Inserted pedestrians and animals are static 3DGS assets, and the \nurec{} render interface accepts only one rigid pose per actor per frame. To keep them from sliding across the road in a frozen posture, we bake limb motion into the geometry as a small bank of copies, each frozen at a different phase of one gait cycle. For pedestrians, we fit the SOMA body model~\citep{Saito2026SOMA} to the asset's Gaussians, starting from walking poses and refining with Chamfer distance. We then bind each Gaussian to the skeleton with skinning weights taken from the nearest body surface and pose the body at $K$ evenly spaced frames of a Kimodo~\citep{Rempe2026Kimodo} walking motion. For animals, we use the SMAL model~\citep{Zuffi2017SMAL} with a procedurally generated trot, solving the knee angle at each frame to keep the stance foot planted. Forward travel and heading drift are removed from every phase, so each copy carries only limb motion and the trajectory still comes from the simulator. At render time, every phase is registered as a separate track. Each frame, phase $\lfloor dK/s \rfloor \bmod K$ is placed at the actor pose and the others are moved out of view, where $d$ is the distance the actor has traveled and $s$ is the stride measured from the baked cycle. Selecting phases by distance rather than by time keeps the feet planted at whatever speed the actor policy chooses. Physics, collision checking, and metrics still see a single track per actor, so the gait affects rendering only. Because each phase occupies render-server GPU memory, the benchmark uses three phases for pedestrians and five for animals.

\subsection{Asset Inventory}
\label{app:asset_stats}

\begin{figure}[t]
  \centering
  \includegraphics[width=0.85\textwidth]{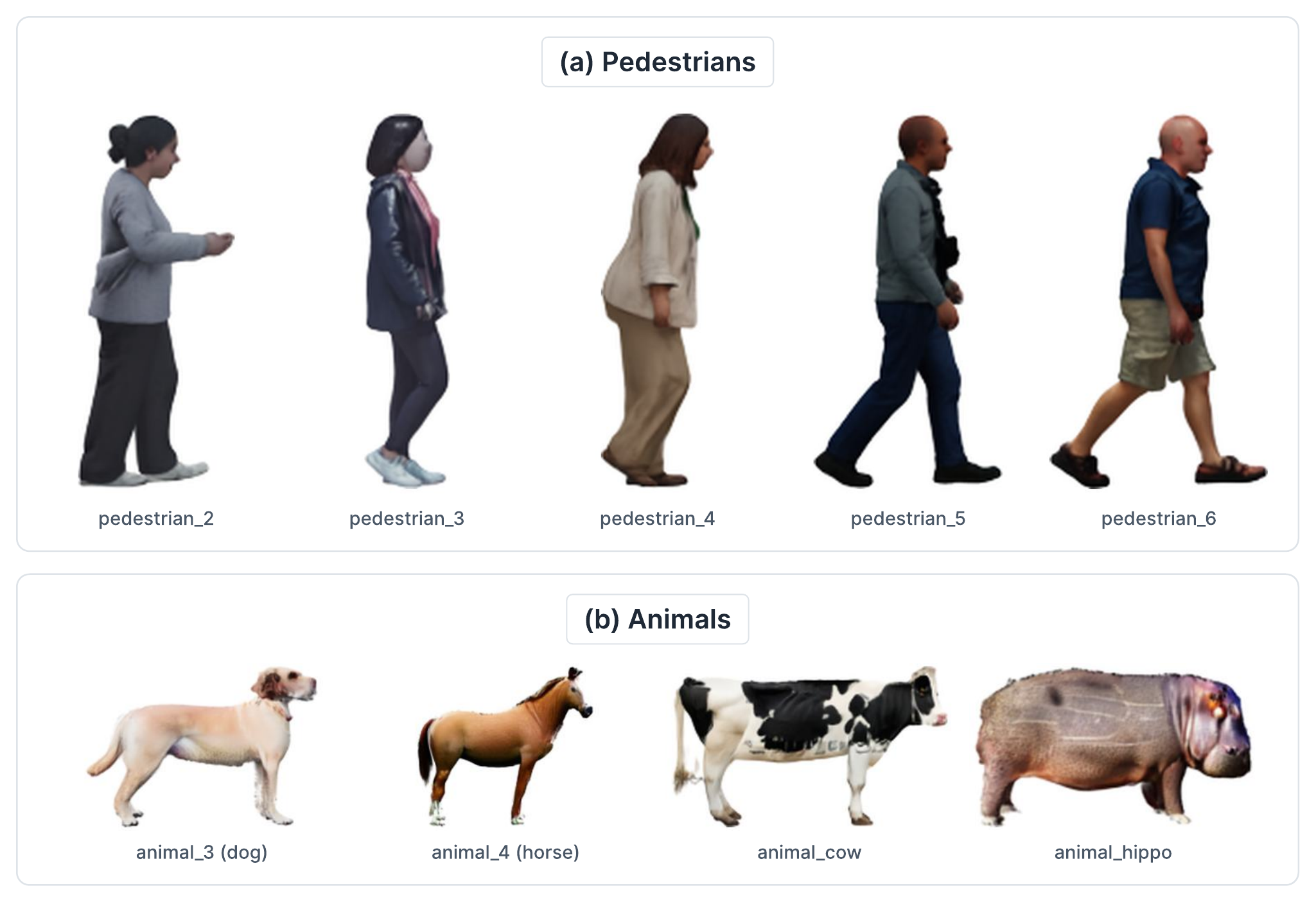}
  \caption{Deformable assets animated with walking gaits: (a) pedestrians and (b) animals.}
  \label{fig:asset_gallery_deformable}
\end{figure}

\begin{figure}[t]
  \centering
  \includegraphics[width=0.85\textwidth]{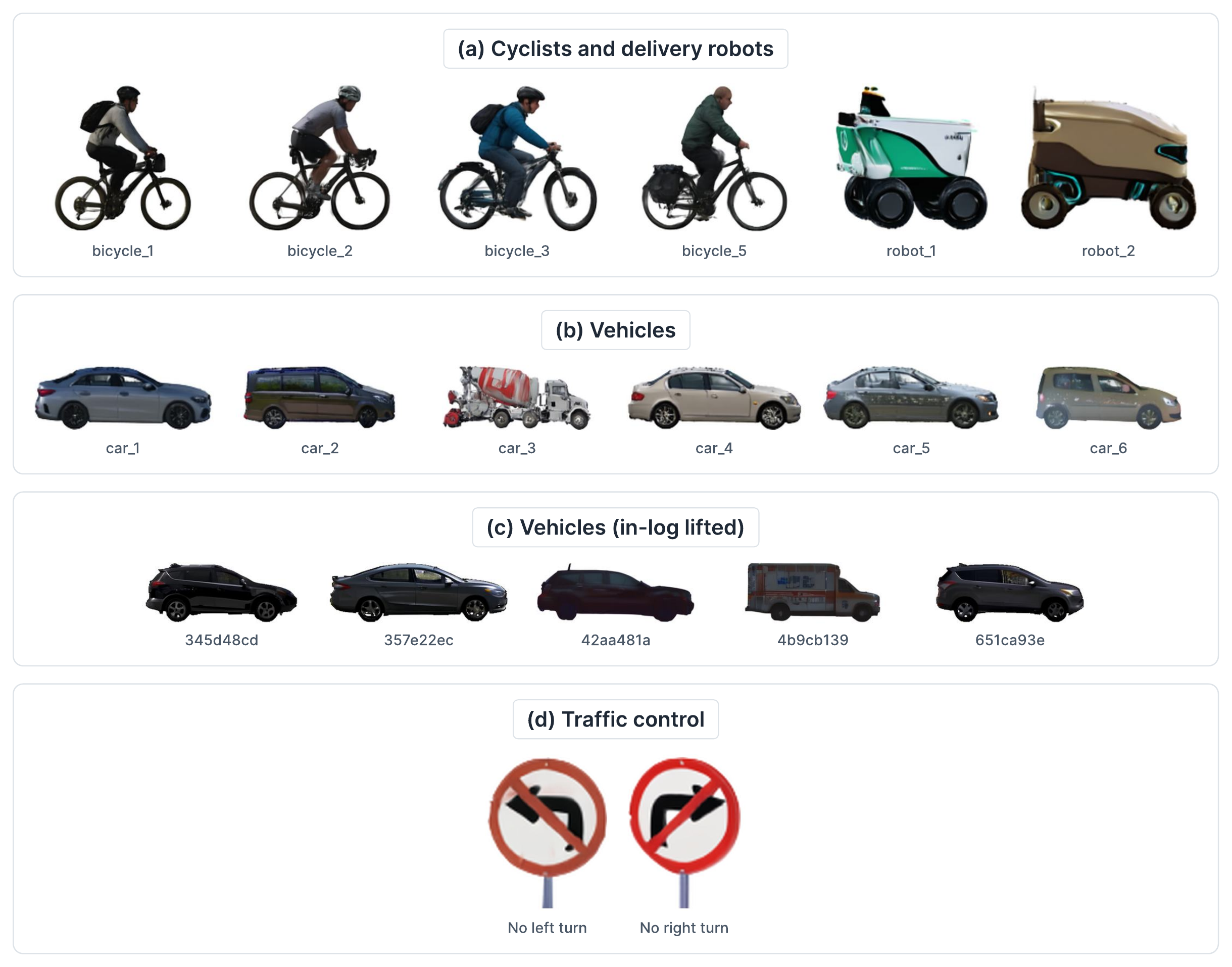}
  \caption{Rigid assets: (a) cyclists and delivery robots, (b) vehicles converted from single nuPlan frames, (c) vehicles converted from the vehicle tracks of the source logs (five examples), and (d) prohibitory signs.}
  \label{fig:asset_gallery_rigid}
\end{figure}

Table~\ref{tab:asset_inventory} lists the asset library by category, source, and the event types that use it. Figures~\ref{fig:asset_gallery_deformable} and~\ref{fig:asset_gallery_rigid} show the assets rendered in isolation.

\begin{table}[H]
\centering
\caption{\navsafe{} asset inventory. nuPlan (track): converted from the vehicle tracks of the source logs; nuPlan (frame): converted from single nuPlan camera frames; Photo: converted from photographs taken by the authors.}
\label{tab:asset_inventory}
\scriptsize
\setlength{\tabcolsep}{4pt}
\renewcommand{\arraystretch}{1.12}
\begin{tabular}{@{}lcll@{}}
\toprule
\textbf{Category} & \textbf{\# Assets} & \textbf{Source} & \textbf{Used by event types} \\
\midrule
Pedestrians & 5 & nuPlan (frame) & R-3 \\
Cyclists & 4 & nuPlan (frame) & R-2 \\
Delivery robots & 2 & nuPlan (frame) & R-3 \\
Animals & 4 & Photo & R-4 \\
Vehicles & 6 & nuPlan (frame) & C-7, V-10, I-3 \\
Vehicles & up to 10 per scenario & nuPlan (track) & All \\
Prohibitory signs & 2 & Photo & V-8 \\
\bottomrule
\end{tabular}
\end{table}

\section{Dataset Source and Photorealistic Reconstruction}
\label{app:environment_details}

This appendix supplements Section~\ref{sec:rendering_fidelity}. It describes how each scenario is reconstructed and rendered, and how the fidelity comparison in Table~\ref{tab:rendering_ablation} is conducted.

\subsection{Source Logs and Evaluation Bundles}
\label{app:source_logs}

Each scenario is a 20\,s window from the nuPlan test split~\citep{Caesar2021nuPlan}. Windows are drawn only from the 147 test logs that provide full-rate camera images, since the remaining test logs contain 2\,Hz images that are too sparse for reconstruction. For every window, the manifest stores the nuPlan scenario token, source log and map, start and end timestamps, event type, recipe hash, and whether an edit is active. Each scenario is packaged as a self-contained bundle containing synchronized ego state, tracked boxes, traffic-light states, route and map context, reconstruction metadata, and referenced assets. Simulation and scoring read this symbolic state, whereas the policy receives only rendered sensor data and navigation inputs.

\subsection{Four-Segment NuRec Reconstruction}
\label{app:reconstruction}

Each scenario is reconstructed as four independently trained 5\,s NuRec models~\citep{NVIDIA2026NuRec,Huang2026InstantNuRec} rather than as one model over the full window: a single 20\,s model spreads the same Gaussian budget over four times as many frames and a correspondingly larger scene extent, which in our preliminary experiments visibly blurred static structure. Reconstruction uses the NRE 26.04 container (\texttt{\seqsplit{nvcr.io/nvidia/nre/nre-ga@sha256:6e0caa70a9148490552520c3dde9ee665c8d094ca10b814601cb8bb306567c90}}) with the \texttt{car2sim\_6cam} configuration. Each segment is trained on five calibrated nuPlan cameras (F0, B0, L0, L1, and R0), using every fourth frame. The top LiDAR initializes the road Gaussians but provides no training rays, since LiDAR supervision yielded only a marginal improvement in visual fidelity at a substantially longer training time. In addition to the background, road, and dynamic-object layers of \texttt{car2sim\_6cam}, we add a layer for static boxed objects (parked vehicles, cones, barriers, and signs), so that every annotated object, moving or static, is a separate node that can be moved, removed, or replaced at render time. To support novel-view rendering of moving vehicles, up to ten of the nearest valid in-log vehicles per scenario are further replaced with assets from the \navsafe{} asset library (Appendix~\ref{app:assets}). Each segment is optimized for 160k iterations with at most 5M Gaussians per segment; Gaussian addition and relocation stop at 140k iterations, and the learning rate follows a step-wise cosine annealing schedule over all 160k iterations.

At render time, each segment is placed in the shared log frame by a stored translation from its local reconstruction frame. Because a segment covers only the road that the logged ego traverses within its 5\,s window, segments are selected by position rather than by simulation time: at each frame, the ego position is matched to the nearest pose on the logged ego trajectory, and the timestamp of that pose determines both the segment to render (the one whose window contains it) and the time at which it is queried. The ego thus enters the next segment when it reaches the corresponding location on the road, regardless of its speed relative to the logged ego, and the switch is made without blending. Because boundaries follow the log rather than event timing, a switch may fall within an event window; such switches are not excluded from evaluation.

\subsection{Rendering and Harmonizer}
\label{app:rendering}

The renderer serves all four segments and renders each camera at its native resolution ($1920\times1080$). Because each scene is a 3DGS reconstruction, any camera rig $\kappa$ can be rendered by changing its intrinsics and extrinsics (Section~\ref{sec:event_design}), including cameras that were not used for training. Each policy adapter selects, crops, resizes, and normalizes the cameras used by its policy: ReCogDrive, MTDrive, SimWAM, DriveLaW, and DriveVLA--W0 use F0; LTF, DiffusionDrive, SparseDriveV2, and GTRS use F0, L0, and R0; DrivoR and RAP use F0, L0, R0, and B0; and AutoVLA uses F0, L1, and R1. DiffusionHarmonizer~\citep{Zhang2026DiffusionHarmonizer} post-processes every rendered frame independently, without temporal conditioning, to reduce reconstruction artifacts and the appearance mismatch of inserted assets.

\subsection{Rendering-Fidelity Evaluation}
\label{app:rendering_fidelity}

\textbf{Evaluation set.} All renderers are evaluated on the front camera (\texttt{CAM\_F0}) over 392 reconstructed 20\,s scenarios, which include the 280 benchmark scenarios, and each rendered frame is compared with the recorded frame of the same timestamp. FID and FVD use the 15{,}680 frames sampled at 2\,Hz, and $\mathrm{FD}_{\pi}$ uses the 4{,}163 NAVSIM \texttt{navtest} tokens~\citep{Dauner2024NAVSIM} among them, since each policy requires a complete NAVSIM scene as input. \navsafe{} renders along the logged ego trajectory at $1920\times1080$ with the Harmonizer enabled, as in closed-loop evaluation. DriveArena~\citep{Yang2025DriveArena} generates each frame at $400\times224$ with its WorldDreamer checkpoint, conditioned on the BEV layout, a text caption, and the previously generated frame; the reference frame is reset at the start of each scenario, and the three logs in DriveArena's training split are excluded. DreamStream~\citep{leng2026dreamstream} generates frames at $1280\times704$ with its Wan~2.2~\citep{Wan2025Wan} (5B) autoregressive video model, conditioned on the logged first frame, per-frame traffic layouts (the projected HD map and 3D agent boxes), and a text prompt. FVD uses the 1{,}568 clips formed by these frames.

\textbf{Metrics.} FID~\citep{Heusel2017FID} uses Inception-v3~\citep{Szegedy2016Inception} features (clean-fid~\citep{Parmar2022CleanFID}) on individual frames, with the recorded frames resized to the resolution of each renderer: $1920\times1080$ for \navsafe{} and DreamStream, whose frames are upsampled from $1280\times704$, and $400\times224$ for DriveArena. FVD~\citep{Unterthiner2018FVD} uses I3D~\citep{Carreira2017I3D} features on clips of 16 consecutive 2\,Hz frames with a stride of 8, resized to $224\times224$. For policy $k$ with scene-context encoder $\phi_k$, let $(\mu_r,\Sigma_r)$ and $(\mu_g,\Sigma_g)$ be the mean and covariance of $\phi_k$ features on recorded and rendered frames. Following \citet{leng2026dreamstream},
\[
\mathrm{FD}_{\pi}^{k}
=
\frac{\lVert\mu_r-\mu_g\rVert_2^2+\operatorname{Tr}\!\big(\Sigma_r+\Sigma_g-2(\Sigma_r\Sigma_g)^{1/2}\big)}{\operatorname{Tr}(\Sigma_r)},
\qquad
\mathrm{FD}_{\pi}
=
\frac{1}{K}\sum_{k=1}^{K}\mathrm{FD}_{\pi}^{k},
\]
with $K=5$ policies (DrivoR, DiffusionDrive, LTF, RAP, and SparseDriveV2). $\phi_k$ is the scene-context feature that each policy passes to its planning head, and Table~\ref{tab:rendering_ablation} reports $\mathrm{FD}_{\pi}$ multiplied by 100. Because FID and FVD depend on the sample size, all renderers are scored on the same frames and clips.

\section{Evaluation Protocol and Metrics}
\label{app:evaluation_protocol}

This appendix supplements Sections~\ref{sec:event_design} and~\ref{sec:protocol}. It specifies how an episode is executed, when it terminates, and how its trace is scored.

\subsection{Closed-Loop Protocol}
\label{app:closed_loop}

\textbf{Timing and warm-up.} The simulator advances at 10\,Hz. Each episode begins with a log-replay warm-up of 20 frames (2\,s) for direct scenarios and 8 frames (0.8\,s) for recipe scenarios, after which every frame is closed loop. The shorter warm-up of recipe scenarios is fixed by the recipe, because the placement of each inserted actor is solved relative to the hand-off frame. Every policy receives the same warm-up on a given scenario.

\textbf{Ego control.} The policy replans every five frames (2\,Hz). A shared tuned-LQR lateral controller with a kinematic bicycle model executes each predicted trajectory, so all policies are evaluated through the same plant.

\textbf{Semi-reactive traffic.} Surrounding traffic runs in a \emph{semi-reactive} mode ported from MetaDrive's reactive scenario traffic~\citep{li2022metadrive}. Every vehicle starts in log replay, and a moving vehicle is handed, one-way, to the Intelligent Driver Model (IDM)~\citep{Treiber2000IDM} once it is oriented within 90° of the ego, lies more than 1\,m behind and within 15\,m laterally of the ego, and has at least 5\,m of logged track remaining. Unlike MetaDrive, which tests these conditions only when a vehicle spawns, we re-test them every frame, so a vehicle that the ego overtakes becomes reactive once it falls behind. Once a vehicle's logged track is exhausted, these conditions no longer apply and it is handed to IDM unconditionally. Only following vehicles are therefore made reactive: vehicles ahead of the ego, pedestrians, cyclists, and static objects replay their logs, and recipe-controlled actors retain their authored behaviors. IDM uses MetaDrive's constants (10\,m minimum gap, 1.5\,s time headway, and 20\,m lead-vehicle search range), evaluated in MetaDrive's km/h speed convention. 

\textbf{Seeds.} Each policy is evaluated on all 280 scenarios under seeds 0, 1,
  and 1024. The seed initializes the Python, NumPy, and PyTorch random number
  generators before the environment and the policy are constructed. Because the
  simulator retains minor nondeterminism, results can differ across seeds even
  for deterministic policies.

\subsection{Event Contracts and Termination Rules}
\label{app:event_contracts}

For route-reaching tasks, the goal is reached when the ego is simultaneously within a 10 m radius of the target and has at most $\max(2\,\mathrm{m}, 0.01L_i)$ of route arc remaining, where $L_i$ is the route length of scenario $i$; arc progress is tracked by a monotone cursor, so leaving and rejoining the route does not retro-credit un-driven stretches. The same predicate is evaluated live and post hoc, so the two cannot disagree. An episode ends at its first terminal event: any contact, 0.5\,s continuously outside the drivable surface, more than 6 m of wrong-way travel, running a red light, an illegal turn, or 5\,s below 0.1\,m/s (deadlock). The deadlock counter is suspended while the ego holds lawfully at a signal or stop line. Following NAVSIM~\citep{Dauner2024NAVSIM}, a collision is charged to the ego only if the ego is moving and the other agent lies ahead of it; rear-end contacts by following agents and contacts while the ego is stationary terminate the episode without being charged. Every episode has a time budget of 60\,s; an episode that reaches neither the goal nor another terminal event ends when the budget expires. Because the episode ends at the goal, and an ego that strays far from the logged path typically leaves the drivable surface and is terminated, rollouts remain close to the reconstructed region. Episodes interrupted by simulator or renderer failures are rerun rather than scored as policy failures.

For most event types, the contract uses the termination rule above, measures Event Progress (EP) as monotone progress along the logged route (100 on goal completion), and counts an episode as successful when it reaches the goal with no evaluated infraction. Four of the 28 event types modify this contract, listed in Table~\ref{tab:leaf_event_contracts}; V-1 appears as two rows because its traffic-light and stop-sign precursors take different contracts. 

\begin{table}[H]
\centering
\caption{Event types whose contract differs from the default.}
\label{tab:leaf_event_contracts}
\scriptsize
\setlength{\tabcolsep}{3pt}
\renewcommand{\arraystretch}{1.12}
\begin{tabular}{@{}p{0.15\textwidth}p{0.80\textwidth}@{}}
\toprule
\textbf{Event type} & \textbf{Modification} \\
\midrule
C-3 Sideswipe & When the ego is assigned a lane change, success additionally requires a lane transition with at least 0.5\,m lateral offset held for five frames. \\
V-1 Red-Light & The goal and deadlock terminals are suspended. Crossing the stop line fails the episode; holding for the full window succeeds. EP is the fraction of the window held before crossing. \\
V-1 Stop-Sign & Success additionally requires 3\,s below 0.3\,m/s before the stop line when the approach is observed. \\
 V-8 Illegal Turn & The goal terminal is suspended and the outside-route-lanes infraction is waived, since the logged route \emph{is} the prohibited turn and a compliant policy must leave it. Entering the prohibited exit lane fails the episode; clearing the intersection without entering the prohibited lane succeeds. \\
I-2 Work Zone & Wrong-way travel is neither a terminal nor an infraction, since the prescribed bypass uses the opposing lane. \\
\bottomrule
\end{tabular}
\end{table}

\subsection{Metrics and Aggregation}
\label{app:evaluation_details}

All metrics are computed from the saved closed-loop trace. For scenario $i$ and seed $s$,
\[
\mathrm{DS}_{i,s}
=\mathrm{EP}_{i,s}\,
(0.50)^{n_{\mathrm{ped}}}
(0.60)^{n_{\mathrm{veh}}}
(0.65)^{n_{\mathrm{layout}}}
(0.70)^{n_{\mathrm{red}}+n_{\mathrm{timeout}}+n_{\mathrm{yield}}}
(0.80)^{n_{\mathrm{stop}}}
\left(1-\frac{p_{\mathrm{outside}}}{100}\right),
\]
where $n_{\mathrm{ped}}$, $n_{\mathrm{veh}}$, and $n_{\mathrm{layout}}$ count at-fault collisions with pedestrians, vehicles, and static objects; $n_{\mathrm{red}}$, $n_{\mathrm{timeout}}$, and $n_{\mathrm{yield}}$ count red-light violations, scenario timeouts, and failures to yield to an emergency vehicle; $n_{\mathrm{stop}}$ counts stop-sign violations; and $p_{\mathrm{outside}}$ is the percentage of driven distance outside the drivable area, excluding the first 0.5\,m of each excursion. The coefficients follow CARLA~\citep{Dosovitskiy2017CARLA} Leaderboard~2.0 and Bench2Drive~\citep{Jia2024Bench2Drive}. An episode succeeds ($S_{i,s}=1$) if it satisfies its event contract: it completes the event and incurs no infraction other than a minimum-speed infraction. A timeout therefore counts as a failure, as does any nonzero $p_{\mathrm{outside}}$. SR is the percentage of successful episodes, $\mathrm{SR}_s = \frac{100}{N}\sum_{i} S_{i,s}$, over the $N=280$ scenarios.

DE is the mean ratio, in percent, between the ego speed and the mean speed of background vehicles at 20 route checkpoints spaced at 5\% of the route. It is undefined, and excluded from the average, when the ego reaches no checkpoint or the median background speed is below 1.5\,m/s, since the ratio is then dominated by queued traffic. Individual checkpoints whose ratio exceeds 1000\% are additionally discarded before averaging. Comfort is the percentage of non-overlapping 2\,s windows in which the smoothed longitudinal and lateral acceleration, jerk, yaw rate, and yaw acceleration all stay within the nuPlan comfort bounds~\citep{Caesar2021nuPlan}.

DS, SR, and Comfort are averaged over the 280 scenarios, and DE over the scenarios where it is defined. The Ability Score of a category is the unweighted mean of the success rates of its event types. We report the mean and the sample standard deviation of each metric over the three seeds.

\subsection{Benchmark Quality Controls and Scope}
\label{app:benchmark_quality}

Every benchmark cell is solvable under the specified evaluation interface: three participants, each assigned one evaluation seed, collectively complete all 840 scenario--seed cells in the Human Driver Reference (Appendix~\ref{app:human_reference}). All scenarios are drawn from the nuPlan test logs underlying the NAVSIM \texttt{navtest} and \texttt{navhard} splits, which share no logs or scenes with the NAVSIM training split \texttt{navtrain}. Each evaluation records the resolved configuration, seed, state trace, infractions, and terminal reason, together with the checkpoint checksum for policy runs, allowing every reported score to be recomputed from its saved trace. These controls make the evaluation auditable while limiting its conclusions to bounded, taxonomy-defined events within the simulator.

\section{Evaluated Policies and Human Reference}
\label{app:experimental_details}

This appendix supplements Table~\ref{tab:main} with the identity and inputs of each evaluated policy and the construction of the Human Driver Reference.

\subsection{Evaluated Policies and Checkpoints}
\label{app:evaluated_policies}

Table~\ref{tab:evaluated_policies} lists the 20 learned policies, which are described in Section~\ref{sec:related_policies}, and the privileged reference PDM-Closed~\citep{Dauner2023PDM}, a rule-based planner that reads ground-truth simulator state and selects among rolled-out proposals. By how they generate the ego trajectory, the 20 learned policies fall into four families:
\begin{itemize}\setlength{\itemsep}{0pt}\setlength{\parskip}{0pt}
  \item \emph{Regression}, which predicts one or several trajectories from learned queries~\citep{Dauner2024NAVSIM,Kirby2026DrivoR}: LTF, LTF (SimScale), DrivoR, DrivoR (SimScale, 134k), and RAP.
  \item \emph{Vocabulary selection}, which scores a fixed set of candidate trajectories~\citep{Li2025GTRS,Sun2026SparseDriveV2}: SparseDriveV2, GTRS-Dense--V2-99, and GTRS-Dense--V2-99 (SimScale).
  \item \emph{Diffusion or flow matching}, which generates trajectories by iteratively denoising sampled motions~\citep{Wang2025ModalCoupling,Liao2025DiffusionDrive}, including VLA backbones with diffusion or flow-matching action heads~\citep{Zhou2026SpanVLA}: DiffusionDrive, DiffusionDrive (SimScale), DiffusionDrive (BeyondDrive), ReCogDrive--2B--IL, ReCogDrive--2B--RL, DriveLaW, DriveVLA--W0, SimWAM, and SimWAM--RL. DiffusionDrive denoises from 20 anchors obtained by $k$-means clustering of training trajectories, using only two denoising steps, so its outputs remain close to these anchors; Section~\ref{sec:method_diagnosis} therefore treats it as anchor-based.
  \item \emph{Autoregressive token decoding}~\citep{Zhou2025AutoVLA}: AutoVLA, MTDrive--SFT, and MTDrive--mtGRPO.
\end{itemize}
Released weights are evaluated without retraining, and each adapter preserves the policy's declared sensor inputs, normalization, and trajectory parameterization (camera preprocessing in Appendix~\ref{app:rendering}; adapter interface in Appendix~\ref{app:toolbox_scoring}). The evaluation manifest pins the model revision, checkpoint checksum, adapter revision, sensor configuration, inference parameters, and controller configuration of every row. Conclusions about a post-training intervention~\citep{Yang2026PostTraining}, such as RL fine-tuning or demonstration perturbation, are drawn only from checkpoint pairs that share the same architecture and differ in that intervention (e.g., ReCogDrive--2B--IL and ReCogDrive--2B--RL, or DiffusionDrive and DiffusionDrive (SimScale)); other comparisons are descriptive. All checkpoints are also mirrored in the \navsafe{} model zoo.

\begingroup
  \scriptsize
  \setlength{\tabcolsep}{3pt}
  \renewcommand{\arraystretch}{1.12}
  \begin{longtable}{@{}p{0.17\textwidth}p{0.11\textwidth}p{0.11\textwidth}p{0.42\textwidth}p{0.09\textwidth}@{}}
  \caption{Evaluated policies, grouped as in Table~\ref{tab:full_leaderboard}. Inputs list the rendered cameras (F0: front; B0: rear; L0/R0 and L1/R1:
  left/right). The last column gives the NAVSIM version each released checkpoint was trained for. Every post-training pair is version-matched, so an intervention is never confounded with a change of training split. All policies are executed through the same controller at the same replan rate, so the parameterization affects only how far ahead each model plans, not how its
  output is tracked. $^{\dagger}$RAP predicts 10 poses over 5\,s at 2\,Hz and AutoVLA predicts 10 poses; both retain the v1 metric heads.}
  \label{tab:evaluated_policies}\\
  \toprule
  \textbf{Policy} & \textbf{Group} & \textbf{Inputs} & \textbf{Description} & \textbf{NAVSIM} \\
  \midrule
  \endfirsthead
  \toprule
  \textbf{Policy} & \textbf{Group} & \textbf{Inputs} & \textbf{Description} & \textbf{NAVSIM} \\
  \midrule
  \endhead
  \midrule
  \multicolumn{5}{r}{Continued on next page}\\
  \endfoot
  \bottomrule
  \endlastfoot
  PDM-Closed & Privileged & Simulator state & Rule-based planner that scores rolled-out proposals using ground-truth scene state. & Rule-based \\
  LTF & IL & F0/L0/R0 & NAVSIM Latent TransFuser baseline that regresses an ego trajectory from logged demonstrations. & v1 \\
  DiffusionDrive & IL & F0/L0/R0 & Truncated diffusion planner that generates multimodal ego trajectories. & v1 \\
  DrivoR & IL & F0/L0/R0/B0 & Register-based planner trained on logged NAVSIM demonstrations. & v1 \\
  RAP & IL & F0/L0/R0/B0 & Planner trained with rasterized counterfactual recovery and cross-agent views and raster-to-real alignment. &
  v1$^{\dagger}$ \\
  SparseDriveV2 & IL & F0/L0/R0 & Vocabulary-based planner that scores factorized paths and velocity profiles in a coarse-to-fine manner. & v1 \\
  GTRS-Dense--V2-99 & IL & F0/L0/R0 & Dense-vocabulary trajectory scorer; V2-99 denotes its VoVNet-99~\citep{Lee2020CenterMask} image backbone. Scores 16{,}384 candidate
  trajectories with per-metric EPDMS heads. & v2 \\
  ReCogDrive--2B--IL & IL & F0 & 2B-parameter VLA with a diffusion trajectory planner, before reward-based post-training. & v1 \\
  MTDrive--SFT & IL & F0 & Supervised VLA stage used as the initialization for MTDrive's multi-turn reward post-training. & v1 \\
  ReCogDrive--2B--RL & RLFT & F0 & ReCogDrive--2B--IL after DiffGRPO reward-based post-training. & v1 \\
  MTDrive--mtGRPO & RLFT & F0 & MTDrive--SFT after multi-turn group-relative reward optimization on offline scenes. & v1 \\
  SimWAM--RL & RLFT & F0 & SimWAM action expert after flow-based group-relative post-training. & v1 \\
  AutoVLA & RLFT & F0/L1/R1 & VLA with adaptive reasoning and reinforcement fine-tuning; only the final released stage is evaluated. &
  v1$^{\dagger}$ \\
  DiffusionDrive {\scriptsize(SimScale)} & Perturbation & F0/L0/R0 & DiffusionDrive co-trained on real logs and SimScale perturbed scenes with
  pseudo-expert supervision. & v1 \\
  DrivoR {\scriptsize(SimScale, 134k)} & Perturbation & F0/L0/R0/B0 & DrivoR trained on 103k real and 134k SimScale samples for 30 epochs. & v1 \\
  LTF {\scriptsize(SimScale)} & Perturbation & F0/L0/R0 & LTF co-trained on real logs and SimScale perturbed scenes. & v1 \\
  GTRS-Dense--V2-99 {\scriptsize(SimScale)} & Perturbation & F0/L0/R0 & GTRS-Dense co-trained on real logs and SimScale perturbed scenes. & v2 \\
  DiffusionDrive {\scriptsize(BeyondDrive)} & Perturbation & F0/L0/R0 & DiffusionDrive trained with expert-proximate unsafe trajectories as
  negatives under BeyondDrive's repulsive objective. & v1 \\
  DriveLaW & World model & F0 & Diffusion planner conditioned on latent features of a jointly trained video world model. & v1 \\
  DriveVLA--W0 & World model & F0 & Emu3~\citep{Wang2024Emu3} visual-token backbone with the released flow-matching action expert. & v1 \\
  SimWAM & World model & F0 & Action-diffusion policy co-trained with a video expert whose isolated attention mask makes the action head
  independent of future frames, so the video branch is dropped at inference. & v1 \\
  \end{longtable}
\endgroup

\subsection{Human Driver Reference}
\label{app:human_reference}

\begin{figure}[tbh]
  \centering
  \includegraphics[width=0.45\linewidth]{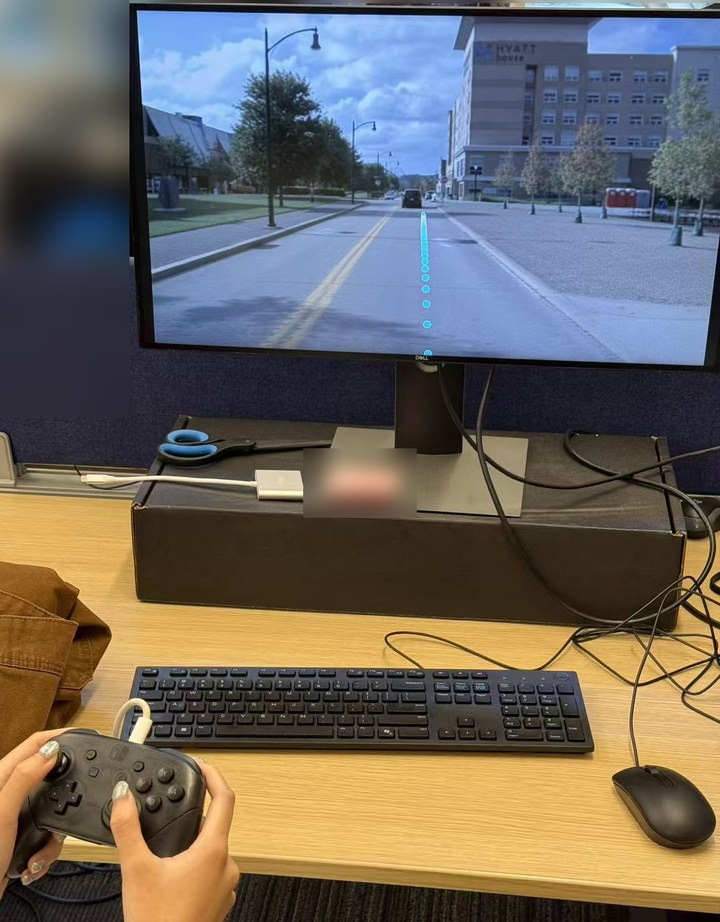}
  \caption{\textbf{Human-reference evaluation setup.}
  The rendered front-camera view with overlaid ground-truth route waypoints and the gamepad used for human-controlled rollouts.}
  \label{fig:human_reference_setup}
\end{figure}

\noindent\textbf{Human participants.}
We recruited three volunteer university students aged 20--30. All participants held valid driver's licenses and reported prior video-gaming experience. The study was conducted under an IRB-approved protocol, and each participant provided written informed consent before participation. Before the recorded evaluation, participants were briefed on the experimental procedure, the rendered front-camera display with overlaid ground-truth route waypoints, and the game-controller interface (Figure~\ref{fig:human_reference_setup}). Each participant then completed a familiarization session in the simulator.

\noindent\textbf{Evaluation procedure.}
Each participant was assigned one of the three evaluation seeds (0, 1, or 1024) and drove all 280 scenarios for that seed, yielding 840 fully human-controlled scenario--seed cells. Participants were instructed to prioritize collision avoidance and compliance with traffic rules while following the displayed route. They observed only the rendered front-camera view and had no access to privileged simulator state, bird's-eye-view observations, future actor trajectories, or the logged ego trajectory. Human commands were issued through a gamepad and executed by the same controller and bicycle model used for policy outputs; vehicle dynamics, traffic evolution, event contracts, termination rules, and scoring were identical to those used in the policy evaluations. The Human Driver Reference establishes scenario solvability under the specified interface rather than population-level human performance or between-driver variability.

\section{Additional Experimental Results}
\label{app:additional_results}

This appendix supplements Section~\ref{sec:experiments} with the OL--CL comparison and complete leaderboard for Section~\ref{sec:leaderboard},
the case-selection protocol for the failure investigations in
Section~\ref{sec:failure_investigations}, and the diagnostic setting for
Section~\ref{sec:method_diagnosis}.

\subsection{Full Leaderboard}
\label{app:full_leaderboard}

Table~\ref{tab:full_leaderboard} extends Table~\ref{tab:main} with three policies omitted for space: SimWAM--RL, DriveLaW, and DriveVLA--W0. All other rows are identical to Table~\ref{tab:main}. Rankings are recomputed over the 20 learned policies.

\begin{table}[t]
  \centering
  \caption{Full \navsafe{} leaderboard, including the world-model-based policies omitted from Table~\ref{tab:main}. Among the 20 learned policies, blue, green, and yellow backgrounds denote the \swone{first-}, \swtwo{second-}, and \swthree{third-}highest point estimates in each column, respectively. Values are means over the same 280 scenarios and seeds 0, 1, and 1024; $\pm$ is the sample standard deviation across the per-seed estimates. Because every event type contributes exactly ten scenarios, per-seed category scores fall on a discrete grid with steps of $10/10$, $10/11$, $10/4$, and $10/3$ points for TC, TV, VRUC, and TI, respectively, and SR equals the event-type-weighted mean of the four ability scores. $^{\mathsection}$ indicates a human controller using \nurec{} front-camera observations; $^{\star}$ denotes a privileged, non-learned reference. Rankings are recomputed over all learned policies; DriveVLA--W0 enters the top three in DE, displacing LTF~{\scriptsize(SimScale)}, and DE is the only column whose highlighting differs from Table~\ref{tab:main}.}
  \label{tab:full_leaderboard}
  \setlength{\tabcolsep}{1.6pt}
  \renewcommand{\arraystretch}{1.15}
  \scriptsize
  \resizebox{\textwidth}{!}{%
  \begin{tabular}{@{}llcccccccc@{}}
      \toprule
      \multirow{2}{*}{\textbf{Policy}} & \multirow{2}{*}{\textbf{Venue}} & \multicolumn{4}{c}{\textbf{Closed-loop metrics}} &
      \multicolumn{4}{c}{\textbf{Closed-loop ability scores}} \\
      \cmidrule(lr){3-6}\cmidrule(l){7-10}
      & & \textbf{DS} & \textbf{SR} & \textbf{DE} & \textbf{Comf.} & \textbf{TC} & \textbf{VRUC} & \textbf{TV} & \textbf{TI} \\
      \midrule
      \textbf{\emph{Human Driver Reference}$^{\mathsection}$} & -- & 100.00{\tiny$\pm$0.00} & 100.00{\tiny$\pm$0.00} & 316.52{\tiny$\pm$0.42} & 75.15{\tiny$\pm$1.89} & 100.00{\tiny$\pm$0.00} & 100.00{\tiny$\pm$0.00} & 100.00{\tiny$\pm$0.00} & 100.00{\tiny$\pm$0.00} \\
      \midrule
      \multicolumn{10}{@{}l}{\emph{Privileged reference}} \\
      PDM-Closed$^{\star}$~\citep{Dauner2023PDM} & CoRL 2023 & 90.17{\tiny$\pm$1.00} & 81.31{\tiny$\pm$1.03} & 305.27{\tiny$\pm$8.57} & 82.49{\tiny$\pm$1.39} & 88.67{\tiny$\pm$1.53} & 72.50{\tiny$\pm$0.00} & 84.24{\tiny$\pm$1.39} & 57.78{\tiny$\pm$1.92} \\
      \midrule
      \multicolumn{10}{@{}l}{\emph{IL-based Methods}} \\
      LTF~\citep{Dauner2024NAVSIM} & NeurIPS 2024 & 53.04{\tiny$\pm$1.92} & 27.74{\tiny$\pm$1.76} & 163.71{\tiny$\pm$0.11} & \cellone{97.81{\tiny$\pm$0.15}} & 31.67{\tiny$\pm$0.58} & 25.00{\tiny$\pm$7.50} & 28.48{\tiny$\pm$1.89} & 15.56{\tiny$\pm$1.92} \\
      DiffusionDrive~\citep{Liao2025DiffusionDrive} & CVPR 2025 & 44.77{\tiny$\pm$0.75} & 20.95{\tiny$\pm$0.55} & 165.08{\tiny$\pm$1.84} & \celltwo{97.68{\tiny$\pm$0.56}} & 21.33{\tiny$\pm$1.15} & 11.67{\tiny$\pm$1.44} & 23.64{\tiny$\pm$0.00} & 22.22{\tiny$\pm$1.92} \\
      DrivoR~\citep{Kirby2026DrivoR} & CVPR 2026 & \cellthree{64.74{\tiny$\pm$1.60}} & \celltwo{50.71{\tiny$\pm$0.71}} & 188.02{\tiny$\pm$1.07} & 87.92{\tiny$\pm$1.25} & \celltwo{58.67{\tiny$\pm$2.52}} & 31.67{\tiny$\pm$5.77} & \cellthree{52.42{\tiny$\pm$2.62}} & 43.33{\tiny$\pm$0.00} \\
      RAP~\citep{Feng2026RAP} & ICLR 2026 & 47.70{\tiny$\pm$0.66} & 27.26{\tiny$\pm$0.41} & \cellthree{220.78{\tiny$\pm$3.77}} & 48.38{\tiny$\pm$0.60} & 32.33{\tiny$\pm$1.15} & 14.17{\tiny$\pm$6.29} & 28.79{\tiny$\pm$0.52} & 22.22{\tiny$\pm$3.85} \\
      SparseDriveV2~\citep{Sun2026SparseDriveV2} & ECCV 2026 & \celltwo{65.26{\tiny$\pm$1.72}} & 48.81{\tiny$\pm$3.04} & 167.99{\tiny$\pm$1.01} & 93.41{\tiny$\pm$0.43} & \cellthree{54.00{\tiny$\pm$0.00}} & \celltwo{43.33{\tiny$\pm$8.04}} & 48.18{\tiny$\pm$4.17} & 41.11{\tiny$\pm$3.85} \\
      GTRS-Dense--V2-99~\citep{Li2025GTRS} & arXiv 2025 & 64.27{\tiny$\pm$0.76} & \cellthree{49.52{\tiny$\pm$1.15}} & 163.88{\tiny$\pm$0.87} & 95.90{\tiny$\pm$0.63} & 52.00{\tiny$\pm$1.73} & \cellthree{40.00{\tiny$\pm$2.50}} & \celltwo{54.24{\tiny$\pm$0.52}} & 36.67{\tiny$\pm$0.00} \\
      ReCogDrive--2B--IL~\citep{Li2025ReCogDrive} & ICLR 2026 & 48.98{\tiny$\pm$0.95} & 28.21{\tiny$\pm$1.43} & 166.94{\tiny$\pm$3.15} & \cellthree{97.58{\tiny$\pm$0.11}} & 30.00{\tiny$\pm$1.00} & 16.67{\tiny$\pm$3.82} & 31.21{\tiny$\pm$0.52} & 26.67{\tiny$\pm$3.33} \\
      MTDrive--SFT~\citep{Li2026MTDrive} & arXiv 2026 & 48.78{\tiny$\pm$0.88} & 31.79{\tiny$\pm$1.43} & 173.77{\tiny$\pm$0.76} & 94.51{\tiny$\pm$1.25} & 30.33{\tiny$\pm$1.15} & 15.00{\tiny$\pm$6.61} & 35.76{\tiny$\pm$1.39} & \cellthree{44.44{\tiny$\pm$1.92}} \\
      \midrule
      \multicolumn{10}{@{}l}{\emph{RLFT-based Methods}} \\
      ReCogDrive--2B--RL~\citep{Li2025ReCogDrive} & ICLR 2026 & 53.64{\tiny$\pm$1.42} & 30.95{\tiny$\pm$2.06} & 173.96{\tiny$\pm$1.90} & 85.10{\tiny$\pm$0.49} & 35.33{\tiny$\pm$2.31} & 18.33{\tiny$\pm$6.29} & 32.42{\tiny$\pm$1.39} & 27.78{\tiny$\pm$5.09} \\
      MTDrive--mtGRPO~\citep{Li2026MTDrive} & arXiv 2026 & 55.86{\tiny$\pm$0.50} & 40.60{\tiny$\pm$0.90} & \cellone{277.62{\tiny$\pm$3.17}} & 93.94{\tiny$\pm$0.41} & 40.33{\tiny$\pm$2.89} & 16.67{\tiny$\pm$1.44} & 46.36{\tiny$\pm$0.91} & \cellone{52.22{\tiny$\pm$3.85}} \\
      SimWAM--RL~\citep{Zhao2026SimWAM} & arXiv 2026 & 55.41{\tiny$\pm$1.72} & 29.76{\tiny$\pm$2.86} & 173.08{\tiny$\pm$2.41} & 78.62{\tiny$\pm$0.68} & 35.67{\tiny$\pm$3.06} & 22.50{\tiny$\pm$4.33} & 28.48{\tiny$\pm$1.89} & 24.44{\tiny$\pm$5.09} \\
      AutoVLA~\citep{Zhou2025AutoVLA} & NeurIPS 2025 & 51.05{\tiny$\pm$0.80} & 29.52{\tiny$\pm$2.43} & 186.02{\tiny$\pm$0.58} & 65.37{\tiny$\pm$0.47} & 33.00{\tiny$\pm$2.65} & 15.00{\tiny$\pm$4.33} & 29.70{\tiny$\pm$2.92} & 36.67{\tiny$\pm$8.82} \\
      \midrule
      \multicolumn{10}{@{}l}{\emph{Passive Demonstration Perturbation Methods}} \\
      DiffusionDrive~{\scriptsize(SimScale)}~\citep{Tian2026SimScale} & CVPR 2026 & 52.19{\tiny$\pm$2.05} & 35.24{\tiny$\pm$2.38} & 189.66{\tiny$\pm$4.28} & 96.92{\tiny$\pm$0.34} & 38.33{\tiny$\pm$1.15} & 18.33{\tiny$\pm$7.22} & 36.67{\tiny$\pm$5.01} & 42.22{\tiny$\pm$3.85} \\
      DrivoR~{\scriptsize(SimScale, 134k)}~\citep{Tian2026SimScale} & CVPR 2026 & 57.66{\tiny$\pm$1.32} & 42.50{\tiny$\pm$0.36} & 176.73{\tiny$\pm$0.92} & 85.88{\tiny$\pm$0.39} & 44.00{\tiny$\pm$1.00} & 17.50{\tiny$\pm$2.50} & 48.18{\tiny$\pm$2.41} & \celltwo{50.00{\tiny$\pm$0.00}} \\
      LTF~{\scriptsize(SimScale)}~\citep{Tian2026SimScale} & CVPR 2026 & 52.65{\tiny$\pm$0.88} & 34.05{\tiny$\pm$0.21} & 192.09{\tiny$\pm$0.85} & 97.41{\tiny$\pm$0.48} & 35.00{\tiny$\pm$1.73} & 17.50{\tiny$\pm$2.50} & 38.48{\tiny$\pm$0.52} & 36.67{\tiny$\pm$0.00} \\
      GTRS-Dense--V2-99~{\scriptsize(SimScale)}~\citep{Tian2026SimScale} & CVPR 2026 & \cellone{73.54{\tiny$\pm$0.84}} & \cellone{61.79{\tiny$\pm$1.56}} & 141.50{\tiny$\pm$1.30} & 87.69{\tiny$\pm$1.66} & \cellone{69.33{\tiny$\pm$0.58}} & \cellone{50.00{\tiny$\pm$2.50}} & \cellone{64.24{\tiny$\pm$1.89}} & 43.33{\tiny$\pm$3.33} \\
      DiffusionDrive~{\scriptsize(BeyondDrive)}~\citep{Wang2026BeyondDrive} & ECCV 2026 & 46.61{\tiny$\pm$0.07} & 21.79{\tiny$\pm$0.94} & 172.79{\tiny$\pm$1.26} & 97.24{\tiny$\pm$0.19} & 22.67{\tiny$\pm$1.53} & 10.00{\tiny$\pm$0.00} & 25.15{\tiny$\pm$1.39} & 22.22{\tiny$\pm$1.92} \\
      \midrule
      \multicolumn{10}{@{}l}{\emph{World-Model-based Methods}} \\
      DriveLaW~\citep{Xia2026DriveLaW} & CVPR 2026 & 34.68{\tiny$\pm$0.25} & 1.90{\tiny$\pm$0.21} & 175.50{\tiny$\pm$1.32} & 95.27{\tiny$\pm$0.01} & 1.00{\tiny$\pm$0.00} & 2.50{\tiny$\pm$0.00} & 2.12{\tiny$\pm$0.52} & 3.33{\tiny$\pm$0.00} \\
      DriveVLA--W0~\citep{Li2025DriveVLAW0} & ICLR 2026 & 45.06{\tiny$\pm$0.64} & 27.98{\tiny$\pm$1.80} & \celltwo{256.31{\tiny$\pm$1.01}} & 62.74{\tiny$\pm$1.14} & 26.67{\tiny$\pm$2.31} & 7.50{\tiny$\pm$0.00} & 33.94{\tiny$\pm$2.29} & 37.78{\tiny$\pm$1.92} \\
      SimWAM~\citep{Zhao2026SimWAM} & arXiv 2026 & 57.53{\tiny$\pm$2.72} & 37.50{\tiny$\pm$1.56} & 160.54{\tiny$\pm$1.64} & 97.19{\tiny$\pm$0.27} & 41.33{\tiny$\pm$0.58} & 26.67{\tiny$\pm$7.22} & 40.61{\tiny$\pm$0.52} & 27.78{\tiny$\pm$5.09} \\
      \bottomrule
  \end{tabular}
  }
\end{table}

\subsection{Open-Loop and Closed-Loop Performance}
\label{app:ol_cl_comparison}

\begin{figure}[H]
  \centering
  \includegraphics[width=\textwidth]
  {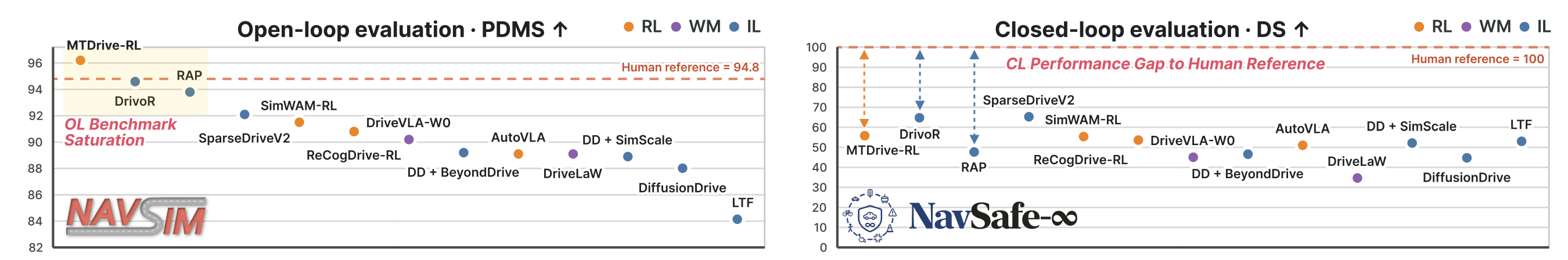}
  \caption{\textbf{Open-loop performance does not determine closed-loop safety.} \textbf{Left:} Published NAVSIM v1 PDMS for the 13 evaluated checkpoints with reported v1 scores. \textbf{Right:} \navsafe{} DS for the same checkpoints under CL execution. Dashed horizontal lines mark the human reference for each benchmark. Several policies approach or exceed the NAVSIM reference but remain substantially below the \navsafe{} reference, and their ordering changes across the two protocols. Colors distinguish IL-only checkpoints (blue),
  RLFT checkpoints (orange), and policies using world-model or model-generated supervision (purple).}
  \label{fig:navsim_navsafe_gap}
\end{figure}

Figure~\ref{fig:navsim_navsafe_gap} compares the 13 evaluated checkpoints
with reported NAVSIM v1 PDMS against their \navsafe{} DS. Because the two scores are defined by different evaluation protocols, their numerical difference is not a calibrated measure of transfer. Instead, the aligned panels show each policy's performance relative to the corresponding human reference.

\subsection{Case Selection for Section~\ref{sec:failure_investigations}}
  \label{app:failure_case_setting}

\textbf{Case selection.} The four diagnostic cases are qualitative illustrations, one per benchmark category, of a failure mode that recurs across that category rather than of a single anomalous episode. Each case was chosen as a representative instance of its category's dominant failure and was fixed before the analysis was carried out; no case was reselected after inspecting its trace. The cases are not a random sample and are not used to estimate how often each failure mode occurs. Every rollout is produced and scored under the benchmark's own protocol---the same termination rules, contracts, and metrics as any other cell (Appendix~\ref{app:event_contracts})---and every panel is regenerated from that cell's saved closed-loop trace. The first three cases use seed~1; the fourth compares two rollouts of one scenario under seeds~1 and~1024. Two cases additionally instrument the policy for inspection without altering what it executes: SimWAM's video expert is re-enabled to decode a future frame, and AutoVLA is run with its chain-of-thought prompt so that a reasoning trace exists at each replan. Both are described below.

\textbf{Planning Failure (LTF, I-3).} The episode ends as an at-fault frontal collision with the incident actor 3.3\,s into the rollout, for a Driving Score of 7.3. The failure is not one of perception: the conflicting actor is present in the policy's own detections well before contact, while the predicted longitudinal profile continues to resemble nominal car following.

\textbf{Scoring Failure (DrivoR, V-1).} The episode ends as an at-fault frontal collision with the lead vehicle 4.9\,s into the rollout, for a Driving Score of 2.6. DrivoR scores its candidate proposals at each replan and executes the highest-ranked one. The rank reported in Section~\ref{sec:failure_investigations} is the position, in that same ordering, of the highest-scoring proposal that is collision-free under the benchmark's own collision check, evaluated at the replan preceding contact; the presence of such a proposal localizes the failure to ranking rather than to candidate generation.

\textbf{Prediction Failure (SimWAM, C-7).} SimWAM discards its video branch at inference by design: an isolated attention mask makes the action expert independent of future frames, so the released policy consumes one frame and emits one trajectory. For this diagnostic only, we re-enable the retained video expert and decode one future frame from the same closed-loop observation. The decoded frame is used solely to inspect the model's predicted world state; it does not enter the action head and does not change the executed trajectory. The episode ends as an at-fault frontal collision with the oncoming vehicle 5.3\,s into the rollout, for a Driving Score of 6.9; the decoded future frame omits that vehicle, and the executed plan passes through the space it occupies.

\textbf{Reasoning Failure (AutoVLA, R-1).} AutoVLA is trained with adaptive reasoning and, under its released prompt, may emit action tokens directly without producing a chain of thought; left to that default it frequently skips reasoning altogether, and there is then no reasoning to compare against the action. For this diagnostic we therefore run the policy with its chain-of-thought system prompt at every replan, so that a reasoning trace is available at each decision point. The prompt is the one released with the checkpoint and is applied identically to both rollouts; the decoded action tokens, and hence the executed trajectory, are read from the same forward pass as the reasoning text. The two policy-induced states are taken from two rollouts of the same scenario under different evaluation seeds, which place the ego in different states by the time the pedestrians are encountered, and are compared at the same simulation timestamp. Both states decode reasoning that calls for a stop, yet only one plan does so: the seed-1 rollout accelerates and ends as an at-fault collision with the crossing pedestrian 7.4\,s into the episode, whereas the seed-1024 rollout stops for the same crossing.

\subsection{Diagnostic Setting for Passive Demonstration Perturbation}
\label{app:perturbation_setting}

The perturbation analysis in Section~\ref{sec:method_diagnosis} (Figure~\ref{fig:augmentation_support}) compares the CL state visitation of the four policies evaluated with SimScale checkpoints in Table~\ref{tab:main} (DiffusionDrive, GTRS-Dense--V2-99, DrivoR, and LTF) before and after SimScale co-training~\citep{Tian2026SimScale}.

\textbf{Scenarios and pairing.} We use two scenarios, R-3 (token \texttt{132307b3c1a55f97}) and V-3 (token \texttt{3a6525cd871a5fff}), drawn at random from the scenarios in which the policy-dependent effect described below occurs. Each policy is evaluated before and after co-training in a paired setting: both rollouts start from the same logged ego pose at $t=0$, route, surrounding actors, and environment configuration, use the same 30 seeds (0--29), and hand control to the policy from $t=0$ without ego replay or additional perturbation. Differences between paired rollouts therefore arise from the checkpoints and their CL behavior. The simulator runs at 10\,Hz, and rollouts are truncated at $t=4$\,s. 

\textbf{Coverage.} Coverage is measured in ego-frame position, which spans the longitudinal and lateral perturbations of SimScale (up to 20\,m and 2\,m). The ego position at $t=2$\,s is expressed relative to the initial pose, $(x_t,y_t)^\top=R(\psi_0)^\top(\mathbf{p}_t-\mathbf{p}_0)$, where $\mathbf{p}_0$ and $\psi_0$ are the ego position and heading at $t=0$, $\mathbf{p}_t$ is the ego position at time $t$, and $R(\psi_0)$ is the 2D rotation matrix by angle $\psi_0$, so that $x_t$ is the forward and $y_t$ the lateral (left-positive) displacement in meters. The background shows the training-state density of the co-trained policies, a weighted mixture of the logged and SimScale state densities: a kernel centered at the logged state at $t=2$\,s and an adaptive kernel density estimate over the SimScale states of the same scene at $t=2$\,s. Colors indicate normalized density, and a rollout state is regarded as covered when it lies in a high-density region of the background.

\textbf{Metrics.} To measure consequences in the benchmark's own units, we reconstruct the cumulative Driving Score on every prefix, $\mathrm{DS}_t=\mathrm{EP}_t\,P_t$, where $\mathrm{EP}_t$ and $P_t$ are the event progress and the penalty product accumulated up to $t$ (Appendix~\ref{app:evaluation_details}), define $r_t=\mathrm{DS}_{t+1}-\mathrm{DS}_t$, and compute the discounted return $G_t=\sum_{k=t}^{T-1}\gamma^{k-t}r_k$ with $\gamma=0.99$ from $t=2$\,s until termination or $t=4$\,s. The reconstructed terminal score matches the evaluated DS for every rollout. Arrows in Figure~\ref{fig:augmentation_support} report $\Delta G=\bar G_{\text{after}}-\bar G_{\text{before}}$. The two-second window isolates the immediate effect of co-training on state visitation; longer horizons add interaction changes and compounding errors that would confound attribution to the augmentation. All results are averaged over seeds.

\textbf{Relation to benchmark results.} The policy-dependent effect is visible across all 280 scenarios (Table~\ref{tab:main}). SimScale co-training raises the DS of the anchor-based DiffusionDrive (44.77$\rightarrow$52.19) and GTRS-Dense--V2-99 (64.27$\rightarrow$73.54), whereas it lowers the DS of the query-based DrivoR (64.74$\rightarrow$57.66) and leaves that of LTF essentially unchanged (53.04$\rightarrow$52.65) despite a higher SR (27.74$\rightarrow$34.05), indicating that the additional successes are offset by lower scores elsewhere. The two scenarios in Figure~\ref{fig:augmentation_support} illustrate the mechanism behind this split.

\subsection{Diagnostic Setting for RLFT}
\label{app:diagnostic_setting}

The RLFT analysis in Section~\ref{sec:method_diagnosis} uses two scenario sets derived from the benchmark: a proxy set for the nominal setting (Figure~\ref{fig:rlft_ol_cl_safety_gap}(b)) and a safety-critical set built on it (Figure~\ref{fig:rlft_ol_cl_safety_gap}(c)).

\textbf{Proxy set.} To reduce evaluation cost, we select one scenario per event type (28 in total) from the benchmark, following the performance-based clustering of tinyBenchmarks~\citep{pmlr-v235-maia-polo24a}. Each scenario is represented by the nominal closed-loop DS and success of 15 development policies. Within each event type, we cluster the scenarios with $k$-means and retain the cluster representative closest to the event-type mean. Five policies excluded from the selection serve as a held-out check: their aggregate DS and SR on the proxy set deviate from those on the full benchmark by 1.58 and 2.43 points on average (mean absolute error). The proxy set supports controlled diagnosis and does not replace the full benchmark. The proxy scenarios are used unmodified, without inserted hazards, as the nominal setting in Figure~\ref{fig:rlft_ol_cl_safety_gap}(b).

\textbf{Safety-critical set.} Each proxy scenario is augmented with two safety-critical events chosen from three types: a vehicle cut-in, lead-vehicle braking, and a vulnerable road user (VRU) crossing the ego path. Event types are assigned according to the road layout and traffic context of each scenario, so that every inserted event is plausible in its host scene; for example, VRU crossings are not inserted on multi-lane arterials with dense, fast-moving traffic and no sidewalks, where pedestrian crossings are unrealistic. The events are authored with the scene-customization workbench of the toolbox (Section~\ref{sec:toolbox}; Appendix~\ref{app:toolbox_authoring}), which places and times each event actor while preserving the route, background traffic, and ego initialization of the host scenario. This set yields Figure~\ref{fig:rlft_ol_cl_safety_gap}(c).

\textbf{Metrics.} CFR is the percentage of episodes without an at-fault collision. MC is the minimum distance between the oriented bounding boxes of the ego and the hazard actor within 4\,s after hazard onset, which is negative when the boxes overlap. In nominal scenarios, which contain no designated hazard actor, MC is the minimum distance to any other actor within 4\,s after scoring begins. Each bar in Figure~\ref{fig:rlft_ol_cl_safety_gap} reports the relative change of the mean metric over matched rollouts, $100\,(\bar M_{\mathrm{RLFT}}/\bar M_{\mathrm{IL/SFT}}-1)$. All results are averaged over three seeds.

\section{Toolbox Implementation}
\label{app:toolbox_details}

This appendix supplements Section~\ref{sec:toolbox} with the implementation of the three toolbox components.

\subsection{Environment Construction}
\label{app:toolbox_environment}

The environment-construction tool converts a driving log into a scenario bundle: four reconstructed segments (\texttt{<token>s1.usdz}--\texttt{<token>s4.usdz}), Arrow state, frame offsets, scenario metadata, assets, and a manifest. Reconstruction runs offline in the pinned NRE container (Appendix~\ref{app:reconstruction}); during a scored run, only rendering, traffic simulation, policy inference, control, and trace recording are performed. The simulator runs on Isaac Sim~\citep{NVIDIA2026IsaacSim} 6.0.0, which hosts the scene and advances the simulation clock; in the evaluated configuration, the ego follows the kinematic bicycle model and images are rendered by the NuRec server. The remaining stack comprises Linux x86-64, Python 3.12, PyTorch 2.10.0 (CUDA 12.8), NumPy 2.3.1, and py123d 0.5.1, resolved by a lock file. Besides nuPlan, the tool supports Waymo Open Dataset logs~\citep{Sun2020Waymo}.

\subsection{Declarative Scenario Authoring}
\label{app:toolbox_authoring}

Every scenario of the nine recipe event types, and every edit, is encoded as a versioned YAML recipe rather than as a simulator-side special case. A recipe specifies the host scenario, frame conventions, ego initialization, actor edits, paired reference, and review metadata. Actor edits support removal, replacement, and insertion. Each external asset carries a content hash and physical dimensions, and each dynamic actor carries a frozen pose track or an explicit controller. The renderer, collision checker, traffic runtime, and scorer resolve the same actor identity and state, so an edit cannot alter the observation without also altering physical interaction. The paired reference disables the edit while retaining the route, background traffic, initialization, and seed. Because a recipe event type is defined entirely through recipes, including scenarios that replay the host log unchanged, all of its scenarios share the same construction and warm-up conventions.

\subsection{Policy Integration, Tracing, and Scoring}
\label{app:toolbox_scoring}

A policy adapter translates the common sensor packet and route context into the native inputs of a checkpoint and converts its output into a common ego-trajectory command. The runtime executes this command through the configured controller and records observations, the requested trajectory, ego and actor states, infractions, traffic transfers, and the terminal reason. Model loading and adapter parameters are resolved from the evaluation manifest rather than stored in the scenario. Because scoring consumes only the saved trace, metric code can be tested independently and results can be re-scored without rerunning inference.

\section{Runtime and Resource Requirements}
\label{app:reproducibility}
\label{app:runtime}

Simulation advances at 10\,Hz, and camera observations are rendered
only at the timesteps where the policy consumes them, so the rendering rate matches the replanning rate (2 Hz here) plus any additional frames needed to populate an observation history. When visualization is enabled, rendering runs
at the full 10\,Hz simulation rate.

We evaluate DrivoR with semi-reactive traffic and the NuRec renderer
on two NVIDIA GeForce RTX 3090 GPUs per worker, one for rendering
and one for policy evaluation. Our evaluation replans at 2\,Hz, so cameras
are rendered at 2\,Hz. Visualization is disabled.
Across 280 scenarios and three random seeds, DrivoR averages
223.68 simulation frames and 206.20 scored frames per episode.
The evaluation loop includes rendering, simulation, policy inference,
control, and scoring. Table~\ref{tab:runtime} summarizes the
resulting runtime and resource requirements.

\begin{table}[H]
\centering
\caption{Runtime and resource requirements
for the DrivoR policy-in-loop evaluation path.}
\label{tab:runtime}
\setlength{\tabcolsep}{3.2pt}
\renewcommand{\arraystretch}{1.12}
\scriptsize
\resizebox{\textwidth}{!}{%
\begin{tabular}{@{}lccccc@{}}
\toprule
\textbf{Quantity} & \textbf{Median} & \textbf{p90} &
\textbf{Maximum} & \textbf{$N$} & \textbf{Conditions} \\
\midrule
Episode evaluation
& 114.0 s & 203.3 s & 271.7 s
& 280 scenarios & $\leq 60$\,s simulated \\
Steady-state frame
& 0.510 s & 0.742 s & 1.032 s
& 280 scenarios & End-to-end \\
Renderer startup
& 50 s & 50 s & 50 s
& 10 launches & Once per worker \\
GPU memory/worker
& 24.1 GiB & 24.7 GiB & 25.1 GiB
& 10 measurements & Summed over 2 GPUs \\
Scenario-bundle storage
& \multicolumn{3}{c}{8.43 GB}
& one bundle & 16-file snapshot \\
\bottomrule
\end{tabular}}
\parbox{\textwidth}{\footnotesize
The renderer starts once per worker and is reused across scenarios.}
\end{table}

\end{document}